\pdfoutput=1

\documentclass[11pt]{article}

\usepackage{ACL2023}

\usepackage{times}
\usepackage{latexsym}
\usepackage{pifont}
\usepackage{newunicodechar}
\usepackage{booktabs}
\usepackage{algorithm}
\usepackage{multirow}
\usepackage{array}
\usepackage{subcaption}
\usepackage{bbm}
\usepackage{enumitem}
\usepackage[noend]{algpseudocode}
\newunicodechar{✓}{\ding{51}}
\newunicodechar{✗}{\ding{55}}

\usepackage[T1]{fontenc}

\usepackage[utf8]{inputenc}
\usepackage{xspace}

\newcommand{\llama}{Expert\xspace}
\newcommand{\llamab}{Base-1B\xspace}
\usepackage{microtype}
\usepackage{amsmath}
\usepackage{graphicx}
\usepackage{amssymb}

\usepackage{inconsolata}
\usepackage[normalem]{ulem}
\useunder{\uline}{\ul}{}

\title{MergeME: Model Merging Techniques \\ for Homogeneous and Heterogeneous MoEs}

\author{Yuhang Zhou$^{\ 1}$\quad Giannis Karamanolakis$^{\ 2}$ \quad Victor Soto$^{\ 2}$ \quad Anna Rumshisky$^{\ 2}$ \quad \\ \textbf{Mayank Kulkarni$^{\ 2}$}\quad \textbf{Furong Huang$^{\ 1}$}\quad \textbf{Wei Ai$^{\ 1}$} \quad \textbf{Jianhua Lu$^{\ 2}$}\\
         $^{1}$ University of Maryland, College Park \\ $^{2}$ Amazon AGI \\ \texttt{\{tonyzhou, aiwei, furongh\}@umd.edu} \quad \texttt{Anna\_Rumshisky@uml.edu} \\
         \texttt{\{karamai, nvmartin, maykul, jianhual\}@amazon.com}}

\begin{document}
\maketitle
\begin{abstract}


The recent success of specialized Large Language Models (LLMs) in domains such as mathematical reasoning and coding has led to growing interest in methods for merging these expert LLMs into a unified Mixture-of-Experts (MoE) model, with the goal of enhancing performance in each domain while retaining effectiveness on general tasks. 
However, the effective merging of expert models remains an open challenge, especially for models with highly divergent weight parameters or different architectures. 
State-of-the-art MoE merging methods only work with homogeneous model architectures and rely on simple unweighted averaging to merge expert layers, which does not address parameter interference and requires extensive fine-tuning of the merged MoE to restore performance. 
To address these limitations, this paper introduces new MoE merging techniques, including strategies to mitigate parameter interference, routing heuristics to reduce the need for MoE fine-tuning, and a novel method for merging experts with different architectures. Extensive experiments across multiple domains demonstrate the effectiveness of our proposed methods, reducing fine-tuning costs, improving performance over state-of-the-art methods, and expanding the applicability of MoE merging.



\end{abstract}

\section{Introduction}
\label{sec:intro}

Large language models (LLMs) pretrained on a wide-variety of corpora have achieved notable success in multiple tasks \cite{touvron2023llama, openai2023gpt4, brown2020language, liu2024large}. With significant progress, there is increasing interest in how to continuously improve the performance of LLMs in new domains, including math \cite{yu2023metamath}, code \cite{roziere2023code}, Wikipedia knowledge \cite{shao2024assisting}, or legal domains \cite{cui2023chatlaw}. One straightforward approach is through continual pretraining (CPT) on domain-specific data, which, however, is challenging for multiple target domains, as it can cause catastrophic forgetting on previously learned tasks~\citep{luo2023empirical}.



An alternative approach is Mixture-of-Experts (MoE) merging, where dense experts are first CPT-ed in parallel for each domain and then merged into a unified MoE model, usually by keeping feedforward neural network (FFN) layers separate and averaging non-FFN layers~\cite{sukhbaatar2024branchtrainmixmixingexpertllms, kang2024self}.
Compared with dense models of similar size, the MoE model uses just a subset of parameters during inference by learning to route tokens to the top few experts, thus reducing inference costs.
Unlike training an MoE model from scratch, MoE merging offers modularity, as individual experts are domain-specialized, and is substantially less expensive, as CPT-ing experts in parallel requires less compute than training the entire MoE on large datasets from the beginning \cite{sukhbaatar2024branchtrainmixmixingexpertllms}.

In this paper, we investigate how to effectively merge different domain expert models into a unified MoE model.
The current state-of-the-art (SoTA) MoE merging approach, such as Branch-Train-Mix (BTX)~\citep{sukhbaatar2024branchtrainmixmixingexpertllms} assumes experts are branched from the same ancestor model and merges experts by simply unweighted averaging the non-FFN layers. 
However, as experts diverge in the parameter space, for example by branching from different ancestors or by training on aggressively different data, unweighted averaging may not effectively handle parameter interference such as sign conflicts~\cite{yu2024language, yadav2024ties}. As a result, the merged MoE may underperform and will require a significant amount of additional fine-tuning to recover in performance, which is both expensive and could be impractical when the experts' training data is not publicly available.
Furthermore, existing MoE merging methods cannot be directly used to merge heterogeneous experts with different architectures, which could be the case in practice, as increasingly more experts are provided by separate teams, such as CodeLlama~\cite{roziere2023code} and Olmo~\cite{groeneveld2024olmo}.
Therefore, it is still an open question how to effectively merge homogeneous and heterogeneous experts into an MoE combining the benefits of each. 


To enable the use of diverse expert models, our work addresses the above limitations via new MoE merging methodologies for both homogeneous and heterogeneous experts. In summary, our work introduces three main contributions:

\begin{itemize}[leftmargin=*]
    \item We utilize advanced merging methods that address parameter interference, demonstrating their superiority over unweighted averaging in homogeneous expert merging, particularly in scenarios with limited resources for post-merging MoE fine-tuning.
    \item We propose a perplexity-based heuristic for routing token sequences to domain-specific experts in low-resource environments where MoE fine-tuning is not feasible.
    \item We develop a novel approach to merge experts with different architectures into a single MoE, which learns to route token sequences dynamically to the appropriate expert. 
\end{itemize}

Through extensive experiments and ablation studies across benchmarks in mathematical reasoning, programming, and general knowledge, we show that our proposed methodologies outperform previous state-of-the-art methods and extend the practical applications of MoE merging.

\section{Background and Related Work}

\subsection{Dense Model Merging}


Dense merging methods combine multiple dense models into one to achieve diverse capabilities \cite{wortsman2022model, ilharco2022editing, goddard2024arcee, jin2022dataless, matena2022merging, roberts2024pretrained}. Most approaches focus on merging homogeneous dense models into another dense model. For example, average merging~\cite{wortsman2022model} averages model parameters, while task vector merging~\cite{ilharco2022editing} adds the unweighted sum of task vectors (the difference between base and expert parameters) back to the dense model with scaling. Other work determines task vector weights instead of using an unweighted sum \cite{jin2022dataless, matena2022merging}. SoTA methods like Dare and Ties \cite{yadav2024ties, yu2024language} trim the task vector to resolve parameter interference: Dare trims the task vector randomly and rescales, while Ties sets vector parameters to zero by magnitude and adjusts signs to reduce conflicts.


In addition to homogeneous model merging, \citet{roberts2024pretrained} propose merging heterogeneous models into a dense model using projectors, while \citet{wan2024knowledge} apply knowledge distillation to fuse heterogeneous models. 
In this work, we introduce a more efficient method for merging experts with limited or no further fine-tuning and, unlike previous work focusing on dense models, we explore merging homogeneous and heterogeneous experts into an MoE model. 

\subsection{MoE Training and Merging}


MoE architectures enable quicker inference with a certain parameter count by introducing Sparse MoE layers, where a router mechanism assigns tokens to the top-$K$ expert FFNs (usually 1 or 2) in parallel \cite{fedus2022switch, shazeer2017outrageously, zhang2022mixtureattentionheadsselecting}. Most MoE training approaches, known as upcycling, train the entire model from scratch to handle multiple tasks \cite{komatsuzaki2022sparse, jiang2024mixtral, dou2024loramoe, dai2024deepseekmoe}. These methods first initialize the MoE model from a pretrained base model and then train it on the entire dataset. However, due to the costly communication between GPUs, the upcycling method introduces significant computational overhead \cite{sukhbaatar2024branchtrainmixmixingexpertllms, li-etal-2024-pedants}. To address this, methods like Branch-Train-Merge (BTM) \cite{gururangan2023scaling, li2022branch} average model outputs from different experts, while Branch-Train-Mix (BTX) \cite{sukhbaatar2024branchtrainmixmixingexpertllms} branches the base model, trains each on different domains, and merges them into a unified MoE. 
BTX is shown to be more effective than BTM as well as dense CPT and MoE upcycling baselines.
Another recent approach, Self-MoE \cite{kang2024self}, uses low-rank adaptation (LoRA) \cite{hu2021lora} to fine-tune experts on generated synthetic data \cite{liu2024csrec} and combines trained adapters into an MoE.
To our knowledge, we are the first to introduce a framework for merging heterogeneous models into an MoE.

\section{Methodology}

\begin{figure}[!t]
     \centering
     \includegraphics[width=\columnwidth]{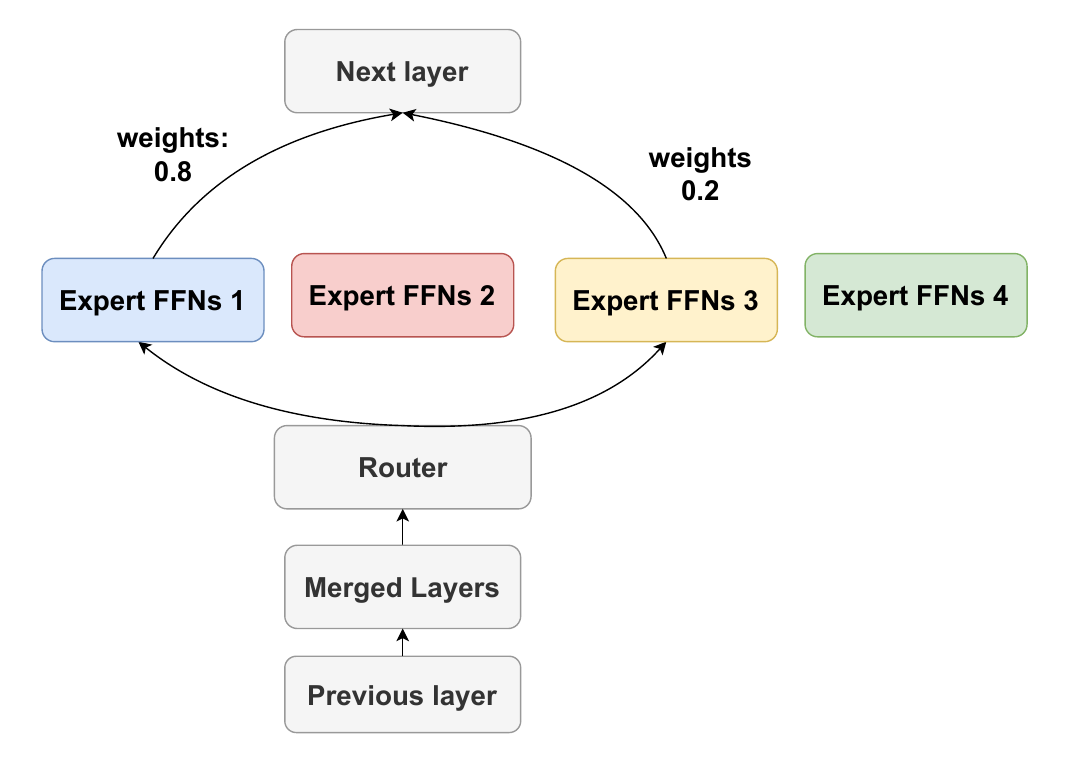}
     \caption{\textbf{Overview of the proposed MoE framework for homogeneous model merging}. We replace averaging with Dare or Ties merging to reduce parameter interference. Additionally, we introduce novel routing heuristics to enhance performance without fine-tuning.}
     \label{fig:moe_homo}
\end{figure}

We define our research problem as follows: Given $l$ dense expert models with parameters $[\theta_1, \theta_2, \dots, \theta_l]$, each pretrained on different domains, we aim to propose an efficient merging method to combine these dense models into an MoE with parameters $\theta_m = \text{Merge}(\theta_1, \theta_2, \dots, \theta_l)$ , optimizing performance across all domains.

We now present our approach for MoE merging with homogeneous and heterogeneous expert models. 
First, for MoE merging with homogeneous experts (Section~\ref{s:homogeneous-model-merging}), we propose replacing existing averaging with more advanced merging methods to deal with parameter interference, and introduce sequence-level routing heuristics to enhance MoE performance without post-merge fine-tuning.
Second, we introduce a novel framework for MoE merging with heterogeneous experts (Section~\ref{sec:merge_with_hetero}), which uses projectors to unify expert inputs and outputs, and a sequence-level router. 

\subsection{Homogeneous Model Merging}
\label{s:homogeneous-model-merging}

First, we describe the basic merging setup (Section~\ref{ss:standard-merging-btx}) and then summarize our extensions to resolve parameter interference (Section~\ref{ss:replacing-average-merging}) and address the need for MoE fine-tuning (Section~\ref{sec:merging_wo_ft}). The overall pipeline is visualized in Figure \ref{fig:moe_homo}.

\subsubsection{Merging Setup}
\label{ss:standard-merging-btx}

Our merging setup is similar to the BTX~\cite{sukhbaatar2024branchtrainmixmixingexpertllms}, where it merges all non-FFN layers (embedding, attention, normalization, and head) of experts by unweighted averaging and keeps the FFNs separate.
As in standard MoE architectures, a router network, implemented as a Multilayer Perceptron (MLP), is inserted between the attention and FFN layers for token-level routing, selecting the top $K$ (usually 1 or 2) experts for each layer among all $l$ experts. The output of FFN layers $\text{FF}_{MoE}(v)$ of token embedding $v$ is formulated as:
\begin{equation*}
    \text{FF}_{MoE}(v) = \sum_{i=1}^K \text{SoftMax}(\text{top-K}(\theta_r v))\text{FF}_i(v)
\end{equation*}
where $\theta_r$ is the parameter of the router network and $\text{FF}_{i}(v)$ is the output of each FFN experts for token $v$.
After merging experts into a single MoE, BTX fine-tunes all parameters, including the router parameters on a mix of training data from all experts.

\subsubsection{Addressing Parameter Interference}
\label{ss:replacing-average-merging}
\begin{figure}[!htb]
     \centering
     \includegraphics[width=\linewidth]{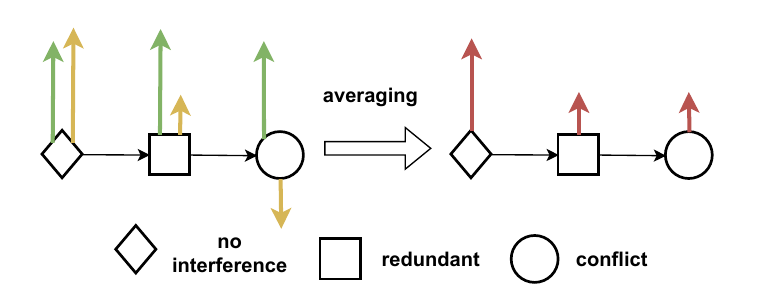}
     \caption{Different types of parameter interference and merged outputs produced by simple averaging.}
     \label{fig:model_interference}
\end{figure}

The major pitfall of the unweighted merging is that there exists parameter interference, as explored in the previous work on dense model merging \cite{yu2024language, yadav2024ties}. As suggested in Figure \ref{fig:model_interference}, when influential parameters (large magnitude parameters) in the task vector merge with redundant parameters (small magnitude parameters) or parameters with sign conflict, simple averaging will output a small magnitude parameter, which may reduce the effect of the original task vector.

In contrast to BTX, we mitigate model interference by employing previous SoTA methods in this MoE setup, namely Dare and Ties. First, we calculate the task vector $\tau_i = \theta_{b} - \theta_i$ with the base model parameter $\theta_b$ and the parameter $\theta_i$ for the model CPTed on domain $i$. 
For Ties merging, we first drop the bottom $(100 - p) \%$ of the redundant parameters (smallest magnitude) by resetting them to 0. For each parameter, we determine the sign with the highest total magnitude in all task vectors and sum all task vectors together to $\tau_m$ but only by keeping the parameter values whose signs are the same as the determined sign.
For Dare merging, we randomly drop the $(100 - p) \%$ parameters. We rescale each task vector with $\tau_i = \frac{\tau_i}{0.01p}$. We sum all task vectors to $\tau_m$.
Finally, we add the summed task vector back to the base model with the scaling term $\lambda$ and obtain the merged layer parameters: $\theta_m = \theta_b + \lambda \cdot \tau_m$. We expect that the drop operation in both methods will address the parameter interference issue, as revealed in dense model merging, and produce a consistent performance boost \cite{yu2024language, yadav2024ties}. 

Similar to BTX, after combining each expert model into an MoE, we fine-tune all parameters in the MoE in the fine-tuning stage. By addressing parameter interference, our approach achieves performance improvements over BTX especially in earlier stages of fine-tuning. Next, we describe how to further reduce the fine-tuning needs.

\subsubsection{Reducing Fine-Tuning Needs}
\label{sec:merging_wo_ft}
Fine-tuning MoEs is expensive due to the communication cost between GPUs \cite{sukhbaatar2024branchtrainmixmixingexpertllms}. Previous MoE merging methods require substantial fine-tuning of the MoE parameters to train the router network. In this section, we propose two techniques to reduce reliance on MoE fine-tuning, namely a perplexity-based routing and separating the attention layers.


The overall MoE pipeline after merging is illustrated in Figure \ref{fig:moe_homo}, but we replace the router network with our routing heuristic to determine the expert selection. Additionally, we separate attention layers without merging them. For each input, the routing heuristic selects the appropriate experts and assigns their weights. The input is then processed by the chosen experts, and their outputs are combined using weights.

\paragraph{Routing Heuristics}


Our goal is to develop routing heuristics that replace the routing network without accessing the training data. We propose a sequence-level heuristics: perplexity (PPL) routing with only access to the inference sentence.

Our approach assesses the confidence of expert models by utilizing perplexity (PPL) to estimate their uncertainty. We then select the experts with the lowest PPL values, indicating higher confidence \cite{jelinek1977perplexity}.
Formally, with the inference input $x_{inf}$ with $t$ tokens and the expert parameter $\theta_i$ for expert $i$, we compute the PPL value $\text{PPL}(x_{inf}, \theta_i)$ as below:
\begin{equation*}
\resizebox{\columnwidth}{!}{%
    $\text{PPL}(x_{inf} \mid \theta_i) = \exp\left(-\frac{1}{t} \sum_{j=1}^{t} \log P(x_j \mid x_{<j}, \theta_i)\right)
    $%
    }
\end{equation*}
where $P(x_j \mid x_{<j}, \theta_i)$ is the probability assigned by model $\theta_i$ on $j$-th token, given previous tokens.

Since a higher PPL indicates greater uncertainty, we use the reciprocal of PPL values to represent the model's confidence. With the top-K routing, the selected experts and their weights $\alpha$ can be computed as follows:

\begin{equation*}
    \resizebox{\columnwidth}{!}{%
        $\alpha = \text{SoftMax}(\text{top-K}(\frac{1}{\text{PPL}(x_{inf} \mid \theta_1)}, \dots, \frac{1}{\text{PPL}(x_{inf} \mid \theta_l)}))$%
        }
\end{equation*}

Additionally, we also propose another routing heuristic based on the task vector and we present the details of this heuristic in Appendix \ref{sec:task_vector_routing}. With the routing heuristics and the corresponding computed weights from the heuristic, we will present the detailed merging process to form the MoE without further fine-tuning.

\paragraph{Separating attention layers}


We hypothesize that by merging attention layers, BTX creates inconsistency between the attention and FFN outputs. Specifically, the merged attention layers are influenced by all $l$ task vectors from the dense experts, while the top-k routing method limits the FFN output to only $k$ task vectors, leading to mismatched outputs.
To address this, we consider keeping experts' attention layers as separate, similar to FFN. This ensures that both the attention and FFN layers come from the same expert, eliminating discrepancies from inconsistent task vector counts.

\subsection{Heterogeneous Model Merging}
\label{sec:merge_with_hetero}
\begin{figure}[!htb]
     \centering
     \includegraphics[width=\linewidth]{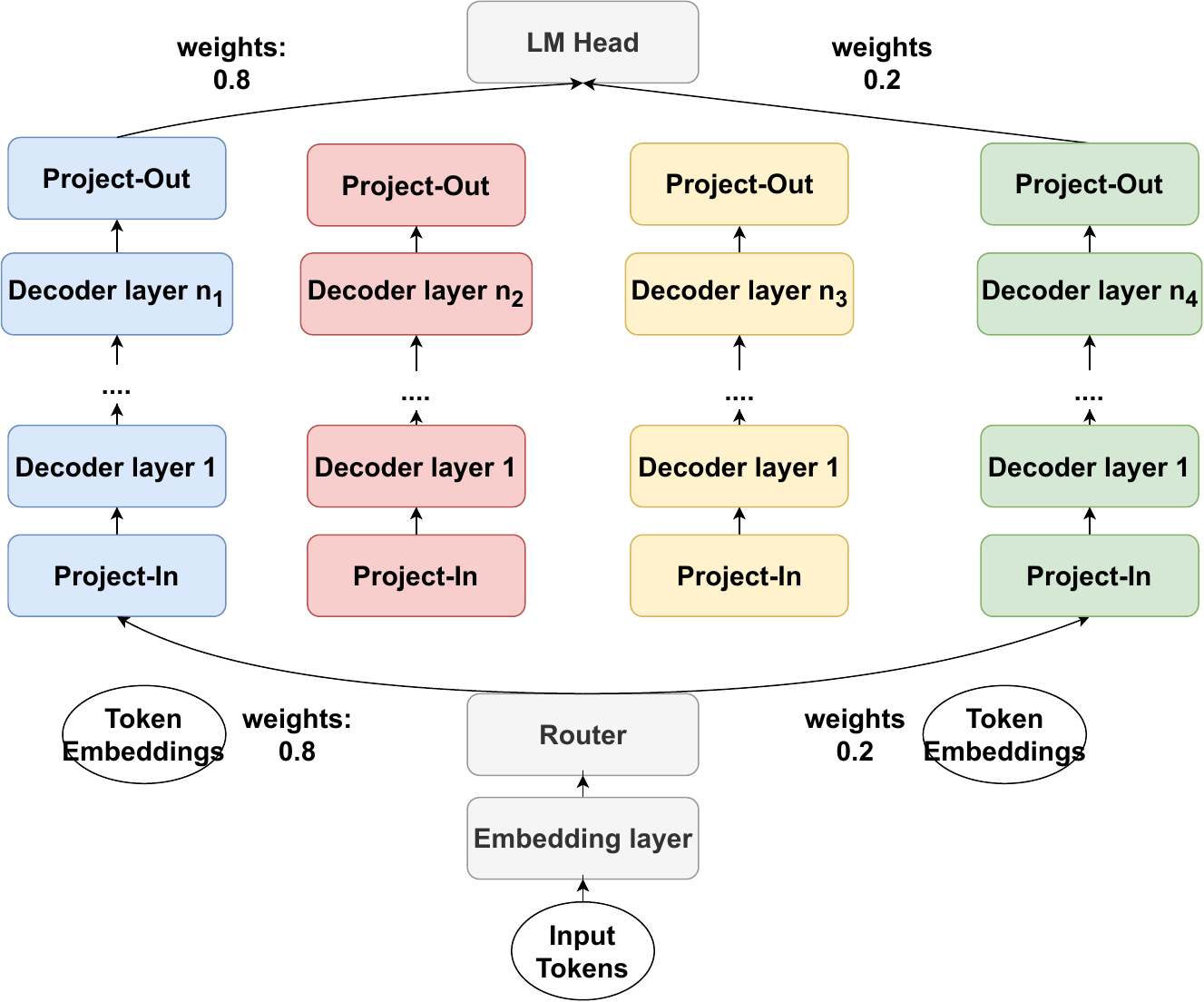}
     \caption{\textbf{Overview of the proposed MoE framework for heterogeneous experts.} Each color represents one heterogeneous expert. $n_1, \cdots, n_4$ refers to the number of layers in each expert.}
     \label{fig:moe_hetero}
\end{figure}

This section describes how to merge models with different architectures into a unified MoE. 
Previous MoE merging techniques cannot be directly used in this setting, as it is not possible to merge non-FFN networks layer by layer when experts have different numbers of layers or different layer shapes. 
To resolve this challenge, we propose a new merging method, which introduces projector layers and sequence-level routing as shown in Figure~\ref{fig:moe_hetero}. 


First, we denote the hidden dimension of all $l$ experts as $d_1, d_2\dots, d_l$, and the maximum dimension among them is $d_m$. Suppose that we have a vocabulary $\mathcal{V}$ and an input sentence with tokens $[v_1, v_2 \dots, v_t]$. For the shared embedding layer $\mathcal{M}_e$, it maps the token $v_i$ in the sentence to embedding $e_i \in \mathbb{R}^{d_m}$ and the shared head layer is the network $\mathcal{M}_{h}: \mathbb{R}^{d_m} \rightarrow \mathbb{R}^{\vert \mathcal{V} \vert}$, which maps the weighted sum of projectors back to the probability distribution of tokens in the vocabulary. The embedding and head layer parameters are initialized from an averaging of the embedding and head layers of each expert. For experts with a hidden dimension less than $d_m$, we add padding zeros for their embedding and head layers before averaging. 


Since we do not merge attention layers due to heterogeneous experts, all tokens must be routed to the same expert. Otherwise, the attention layers cannot perform self-attention, as they require access to every token. Hence, we average the token embeddings and use the router to perform the sequence-level routing. Formally, for top-$K$ routing with router parameters $\theta_r$, the router computes the model weights as follows:
\begin{equation*}
    \alpha = \text{SoftMax}(\text{top-K}(\theta_r \text{avg}(e_1, e_2, \dots, e_t)))
\end{equation*}

For projectors: Proj-in and Proj-out, for each expert, randomly initialized MLP layers, they project the embedding outputs to the dimension of each expert, and project the expert output back to the maximum dimension. For $i$-th expert, we define:
\begin{equation*}
\resizebox{\columnwidth}{!}{%
$
\text{Proj-in layer}: \mathbb{R}^{d_m} \rightarrow \mathbb{R}^{d_i}, \quad \text{Proj-out layer}: \mathbb{R}^{d_i} \rightarrow \mathbb{R}^{d_m}
$
}
\end{equation*}


After using the selected $K$ experts to process the input sequences and translating their outputs to the representation $r_i$ via the Proj-out layer (with dimension $d_m$), we combine the representations using the router's weights: $\sum_{i=1}^K \alpha_i r_i$. The combined representation is then fed into the head layer to obtain the token probabilities. 

After merging the heterogeneous experts into the MoE model, we choose an arbitrary tokenizer from one expert, following previous work \cite{roberts2024pretrained} and fine-tune all parameters.
\section{Experiments Setup and Model Analysis}
\label{sec:experiments}
Through our extensive empirical analysis, we aim to evaluate our frameworks in the settting of homogeneous experts and heterogeneous experts.

\subsection{Evaluation Dataset}
We evaluate our proposed methodology on 6 datasets from three domains, as in the previous work \cite{sukhbaatar2024branchtrainmixmixingexpertllms}. For math reasoning, we choose GSM8K (8-shot) and MATH (4-shot) \cite{cobbe2021training, hendrycks2021measuring}. For code generation, we choose MBPP (0-shot) and HumanEval (0-shot) \cite{chen2021evaluating, austin2021program}. For world knowledge, we choose Natural Questions (NQ, 5-shot) and TriviaQA (5-shot) \cite{kwiatkowski2019natural, joshi2017triviaqa}.

\subsection{Model Configuration}
\label{sec:model_pretraining}
This section describes the base model and experts discussed in our experiments:

\begin{itemize}[leftmargin=*]
\item \textbf{Base Model (\llamab}): This is our base model with 1B parameters and Llama-like architecture. We pretrain \llamab from scratch with 250 billion (250B) tokens from the following datasets from the RedPajama dataset~\cite{together2023redpajama}: Arxiv, CommonCrawl, C4, StackExchange data and the first half of the WikiPedia data in the RedPajama dataset. 
    \item \textbf{Math \llama}: We CPT the Base model on the OpenWebMath data for 100B tokens ~\cite{paster2023openwebmath}.
    \item \textbf{Code \llama}: We use the GitHub data in RedPajama to CPT the Base model for 100B tokens.
    \item \textbf{Knowledge \llama}: We CPT the \llamab model on the second half of the Wikipedia data in the RedPajama dataset for 100B tokens.
    \item \textbf{Math TinyLlama and Math Olmo}: We CPT the TinyLlama-1.1B model \cite{zhang2024tinyllama} and Olmo-1B model \cite{groeneveld2024olmo} on the same data mixture of the Math \llama.
    \item \textbf{Mixture of Experts (MoE)}: 
    For homogeneous model merging, we combine three experts (Math \llama, Code \llama, Knowledge \llama) and one base model (\llamab) into an MoE. For heterogeneous merging, we combine Code \llama, Knowledge \llama, \llamab, and either Math TinyLlama or Math Olmo. MoE fine-tuning is performed on all data sources from the base and expert models, using an additional 40B tokens. Detailed sampling ratios for pretraining and fine-tuning are provided in Appendix \ref{sec:data_mix}.
\end{itemize}

We present the details of model architecture for each expert in Appendix \ref{sec:implement}.


\subsection{Baseline Methods}
\label{sec:baseline}
To demonstrate the effectiveness of our methodology, we compare the performance of the merged 4-expert MoE models with several other baselines.

\begin{itemize}[leftmargin=*]
    \item \textbf{Base \& Experts}: The dense base and expert models in Section \ref{sec:model_pretraining}.
    
    
    
    \item \textbf{BTX} \cite{sukhbaatar2024branchtrainmixmixingexpertllms}: The MoE model derived from the BTX pipeline with average merging and post-merge fine-tuning.
    \item \textbf{Random Routing}: The average merged MoE with randomly initialized router.
    \item \textbf{Router Fine-tuning}: The MoE model derived from the BTX pipeline but only fine-tune the parameters in the router network.
    \item \textbf{3-expert MoE}: 
    To demonstrate the functionality of Math Olmo or TinyLlama in heterogeneous expert merging, we prepare 3-expert MoE models (Base, Knowledge \llama, Code \llama), fine-tuned either on the full data source (including math) or only on code- and knowledge-related data. We merge these models using the BTX method, naming them \textbf{3-expert MoE (same data)} and \textbf{3-expert MoE (w/o math)}.
    \item  \textbf{Dare Dense} \cite{yu2024language}, \textbf{Ties Dense} \cite{yadav2024ties}: Advanced dense model merging method. We apply Dare or Ties to merge four LMs to one dense model.
\end{itemize}

The details of the model configuration of the baseline methods are included in Appendix \ref{sec:implement}.

\subsection{Similarity of Model Parameters}

Before presenting the performance of our proposed methodology, we first analyze the similarities in model parameters across different experts to demonstrate the necessity for alternatives to average merging. Previous work assumes that parameters in attention layers are less domain-specialized, leading to the use of simple averaging when combining non-FFN layers \cite{sukhbaatar2024branchtrainmixmixingexpertllms}. Our analysis aims to verify whether this assumption holds true for experts trained on different domains.

\begin{figure}[!htb]
     \centering
     \includegraphics[width=\linewidth]{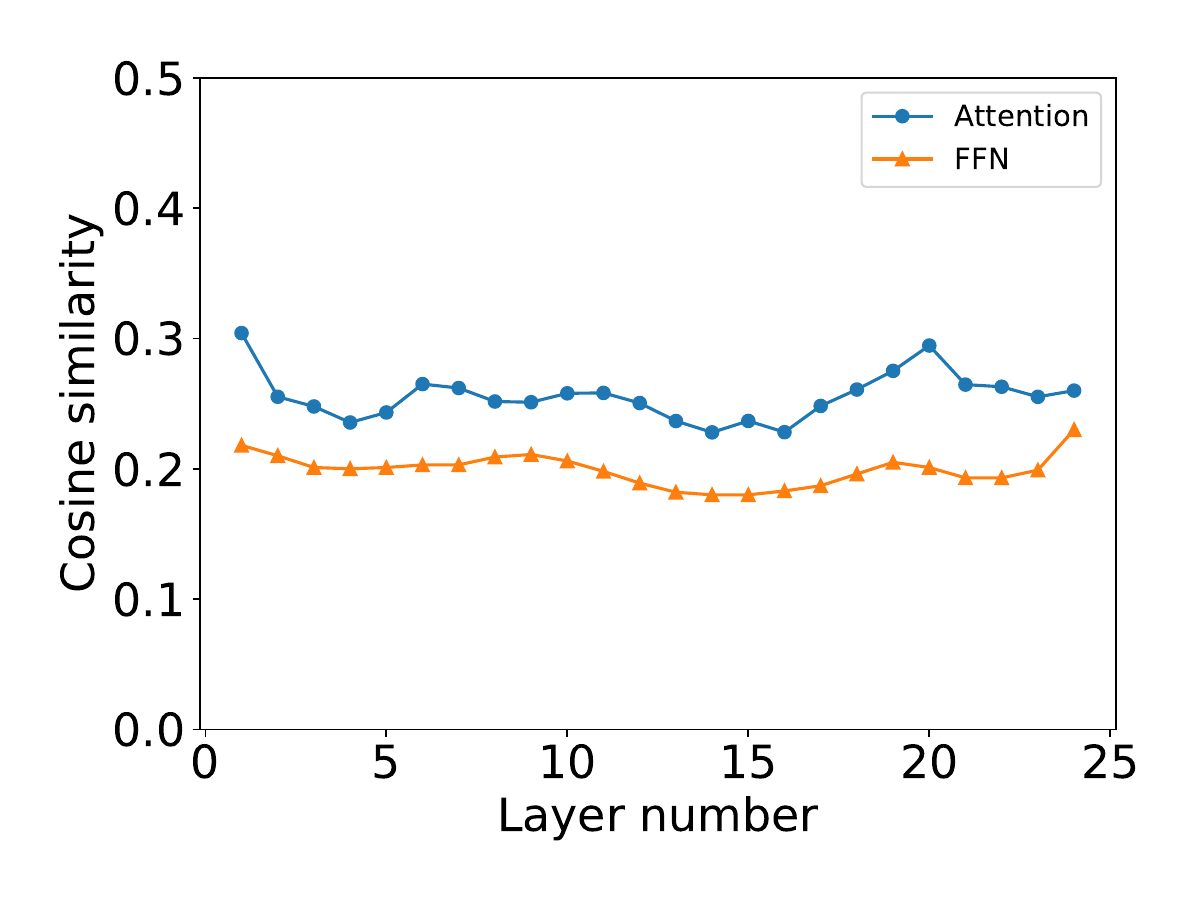}
     \caption{\textbf{Similarity of task vector for attention and FFNs layers for Math and Code \llama experts.} We average the similarity of attentions or FFNs in one decoder layers as the overall similarity for each layer.}
     \label{fig:simi_task_vector}
\end{figure}

To quantify the degree of domain specialization in the model layers, we first extract the task vectors for each layer from our Math and Code \llama models. We then concatenate the task vectors from the attention layers and FFNs into two long vectors. Next, we calculate the cosine similarity between the two concatenated task vectors. The cosine similarity for the task vectors of the FFNs and self-attention layers is visualized separately in Figure \ref{fig:simi_task_vector}.

We observe that the task vectors from both layers exhibit low similarity, suggesting that the assumption of similar attention layers does not consistently hold and parameter interference may occur. This analysis demonstrates the need for more advanced merging methods, rather than averaging, for homogeneous model merging. 




\section{Results}

\subsection{Homogeneous Model Merging}

\subsubsection{Averaging vs. Dare / Ties}
\label{sec:rq1}

\textbf{Replacing simple averaging with Dare or Ties merging obtains better performance.} \quad
In this section, we demonstrate the superiority of our proposed Ties and Dare merging MoE over the BTX merging method.
We present the performance of MoE models with \textbf{Dare merging} or \textbf{Ties merging} on non-FFN layers and other baselines in Table \ref{table:homo_results}. The details of training cost for each method are presented in Table \ref{table:1_training_cost} in Appendix.

\begin{table}[!htb]
\centering
\resizebox{\columnwidth}{!}{%
\begin{tabular}{lccccccc}
\hline
Method             & MBPP           & HumanEval      & MATH          & GSM8K         & NQ            & TriviaQA       & Avg.           \\ \hline
\multicolumn{8}{c}{\textbf{Dense Model}}                                                                                               \\ \hline
\llamab & 4.60            & 3.04           & 2.42          & 1.44          & \textbf{6.61} & 26.72          & 7.47           \\ \hline
Code \llama        & \textbf{10.2}  & \textbf{8.53}  & 2.42          & 2.57          & 3.11          & 16.70           & 7.26           \\ \hline
Math \llama        & 9.80            & 6.71           & \textbf{7.81} & \textbf{6.36} & 5.48          & 19.86          & \textbf{9.34}  \\ \hline
Knowledge \llama   & 3.60            & 4.26           & 2.62          & 2.04          & 5.65          & \textbf{28.71} & 7.81           \\ \hline
\multicolumn{8}{c}{\textbf{MoE Merging}}                                                                                                 \\ \hline
Random Routing     & 4.00           & 6.10           & 2.78          & 2.05          & 4.86          & 21.75          & 6.92           \\ \hline
Router Fine-tuning & 3.60           & 6.71           & 2.42          & 2.96          & 5.82          & 25.98          & 7.92           \\ \hline
BTX merging        & 12.40          & 11.58          & 6.74          & 7.73          & \textbf{6.78} & 25.10           & 11.72          \\ \hline
Ties merging       & 14.20          & \textbf{11.98} & 6.74          & 7.81          & 6.72          & 27.66          & 12.52          \\ \hline
Dare merging       & \textbf{14.20} & 10.98          & \textbf{6.82} & \textbf{7.96} & 6.50           & \textbf{30.68} & \textbf{12.86} \\ \hline
\multicolumn{8}{c}{\textbf{MoE from Scratch}}                                                                                                \\ \hline
MoE Upcycling & 18.40  & 12.20 & 7.80 & 12.21 & 8.37  &  37.33 & 16.05 \\
\hline
\end{tabular}
}
\caption{\label{table:homo_results} \textbf{Performance of proposed Dare and Ties merged MoE and other baselines across six datasets.} The best performance of Dense and MoE model is marked in bold. Results of Dare and Ties merged MoE outperform the BTX MoE and other baseline methods.}
\end{table}


From Table \ref{table:homo_results}, we see that individual experts generally achieve the best performance in their respective domains, as expected. However, CPTed \llama models experience catastrophic forgetting. For instance, both Code and Math \llama perform worse than \llamab on the TriviaQA and NQ datasets.


The results in Table \ref{table:homo_results} show that using Ties or Dare merging significantly improves MoE performance over the BTX pipeline across almost all datasets, with a relative improvement of 6.94\% and 9.72\% in average performance. This suggests that advanced merging methods reduce weight interference and enhance performance.


As a reference, we include the results of MoE sparse upcycling \cite{komatsuzaki2022sparse} in the last row of Table \ref{table:homo_results}. This approach initializes the MoE model by creating four identical copies of the FFN layers from the base model and then CPT on the same 340B tokens used in our pipeline. However, we do not directly compare our results with the upcycling method, as it involves pretraining the entire MoE on all data, incurring significantly higher costs.
We also visualize the average performance for each merging method with different fine-tuning token numbers in Figure \ref{fig:result_token} in Appendix \ref{sec:supp_routing}.
In Figure \ref{fig:result_token}, we observe that the Dare and Ties merging MoE models consistently outperform the BTX merging MoE throughout fine-tuning, especially in the earlier stages of fine-tuning. 

\begin{figure}[!t]
    \centering
    \begin{subfigure}[b]{0.48\columnwidth}
        \centering
        \includegraphics[width=\textwidth]{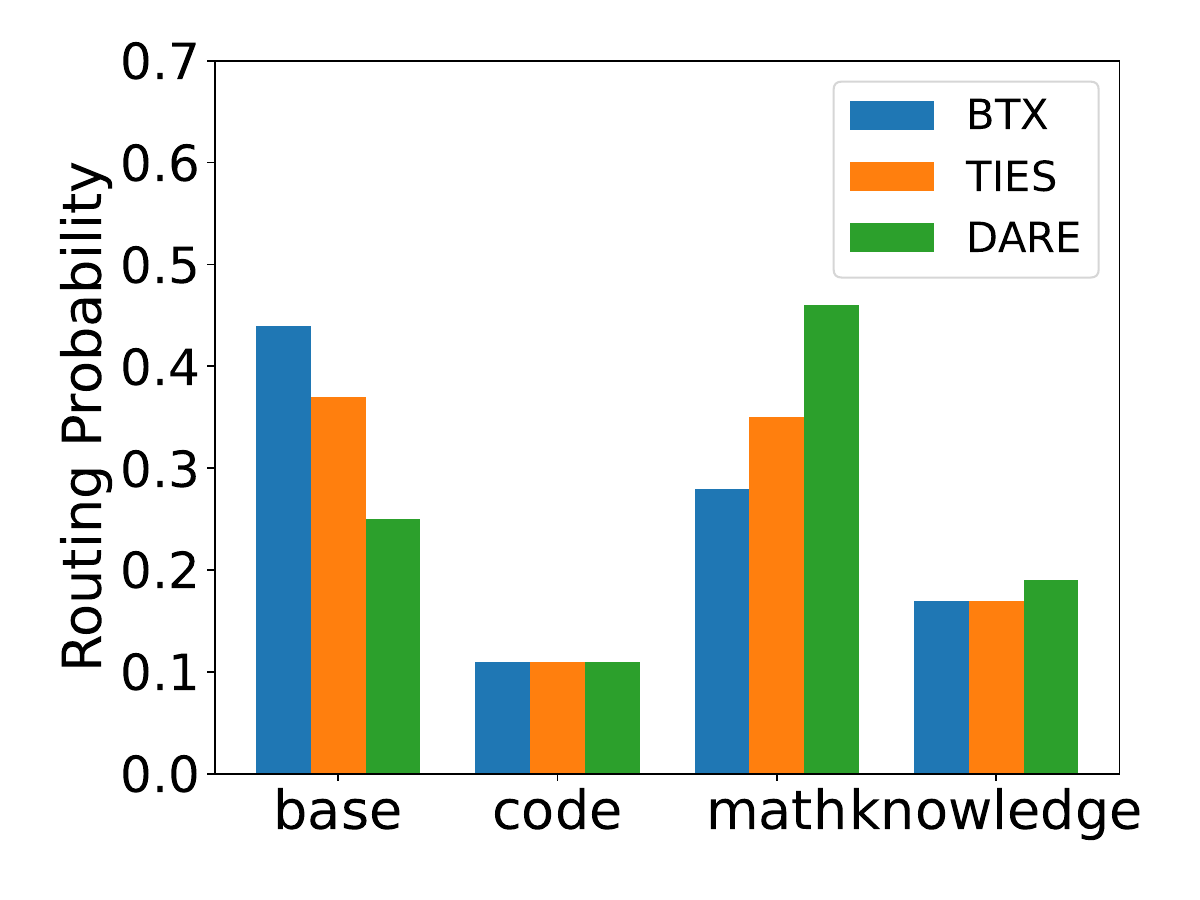}
        \caption{GSM8K}
        \label{fig:gsm8k_route}
    \end{subfigure}
    \hfill
    \begin{subfigure}[b]{0.48\columnwidth}
        \centering
        \includegraphics[width=\textwidth]{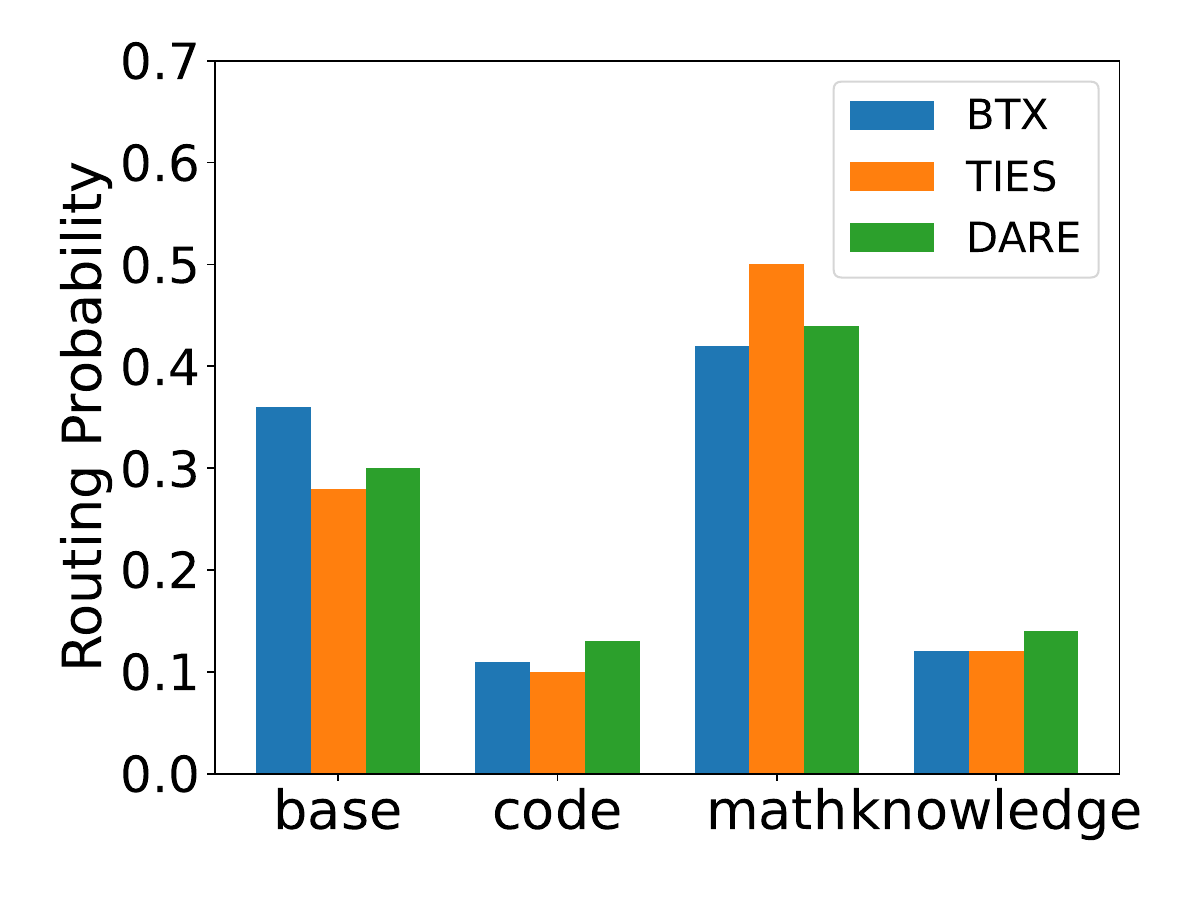}
        \caption{MATH}
        \label{fig:math_route}
    \end{subfigure}
    \vspace{-0.5em}
    \caption{Routing probability of experts on GSM8K and MATH for different merging methods.}
    \label{fig:routing_prob}
\end{figure}

\noindent \textbf{MoE with Dare or Ties merging routes more tokens to domain experts.}\quad
To further explore the effectiveness of Dare and Ties merging MoE, we evaluate MoEs on multiple benchmarks and calculate the routing probability averaged from each layer and token. We visualize the routing probability of each method of two math datasets (MATH and GSM8K) in Figure \ref{fig:routing_prob} and for other datasets, we put the results in Figure \ref{fig:supp_routing_prob} in Appendix \ref{sec:supp_routing}.

Compared to MoEs with BTX merging, where the base model accepts the most routing decisions, the Dare and Ties merging method routes tokens to domain experts more frequently, as suggested in Figure \ref{fig:routing_prob}. For example, for the GSM8K dataset, the routing probability for math expert increases from 0.28 to 0.35 or 0.46 when replacing simple averaging with the Ties or Dare merging. This finding suggests that the more effective MoE with the more advanced merging method should be attributed to more optimized routing decisions.

\subsubsection{Merging without Fine-tuning}

In this part, we will evaluate our proposed routing heuristics in Section \ref{sec:merging_wo_ft} for MoE without fine-tuning.
Before we evaluate the overall performance of each benchmark, we will first examine the routing decision with our proposed heuristics. We present the routing probability for PPL routing heuristics for each dataset in Table \ref{table:routing_heuristic}.

\begin{table}[!htb]
\centering
\resizebox{0.8\columnwidth}{!}{%
\begin{tabular}{lcccc}
\hline
Benchmark      & Base       & Code          & Math          & Knowledge     \\ \hline
GSM8K     & 23\%       & 2\%           & \textbf{43\%} & {\ul 32\%}    \\ \hline
MATH      & 22\%       & 2\%           & \textbf{49\%} & {\ul 27\%}    \\ \hline
MBPP      & 19\%       & {\ul 22\%}    & \textbf{44\%} & 15\%          \\ \hline
HumanEval & 5\%        & {\ul 43\%}    & \textbf{45\%} & 7\%           \\ \hline
NQ        & {\ul 43\%} & 4\%           & 10\%          & \textbf{43\%} \\ \hline
TriviaQA  & {\ul 50\%} & 0\%           & 0\%           & \textbf{50\%} \\ \hline
\end{tabular}
}
\caption{\label{table:routing_heuristic} \textbf{Routing probability of PPL routing for each dataset.} The largest probability are in bold, and the second-largest are underlined.}
\end{table}

\noindent \textbf{Routing heuristic effectively assigns tokens to the corresponding experts.} \quad 
Table \ref{table:routing_heuristic} demonstrates that PPL routing generally achieves the desired routing patterns, effectively directing inputs from a specific domain to the specialized expert models, except in the case of the MBPP dataset. Since our heuristics rely solely on inference inputs without fine-tuning, they can be considered reliable strategies. We also visualize the routing probability for both PPL and task vector routing heuristics for each dataset in Figure \ref{fig:routing_heuristic} in Appendix \ref{sec:supp_routing}. We find that PPL routing consistently produces better results than the task vector routing.

Next, we evaluate the performance on each dataset with different combinations of merging methods and routing heuristics, compared to the baseline methods. We prepare three dense fine-tuning baselines: \textbf{Dare Dense}, \textbf{Ties Dense} and \textbf{Random Routing} (details in Section \ref{sec:baseline}). We also evaluate the ablation methods: merging attention layers without separation and task vector routing. We present the results of each method across datasets in Table \ref{table:moe_wo_finetune}. The details of training cost for each method are presented in Table \ref{table:3_training_cost} in Appendix.

\begin{table}[!htb]
\centering
\resizebox{\columnwidth}{!}{%
\begin{tabular}{llccccccc}
\hline
Merging      & Routing     & MBPP         & HumanEval     & MATH          & GSM8K         & NQ            & TriviaQA       & Avg.          \\ \hline
\multicolumn{9}{c}{\textbf{Dense Merging}}                                                                                                          \\ \hline
Dare       & N/A         & 6.20         & 6.70          & 2.22          & 2.27          & 4.80          & 20.45          & 7.11          \\ \hline
Ties        & N/A         & 6.00         & 6.70          & 2.48          & 2.19          & 3.62          & 20.86          & 6.98          \\ \hline
\multicolumn{9}{c}{\textbf{MoE Merging}}                                                                                                            \\ \hline
Merge attention     & random      & 4.00         & 6.10          & 2.78          & 2.05          & 4.86          & 21.75          & 6.92          \\ \hline
Merge attention     & task vector & 6.60          & 4.87          & 3.06          & 1.44          & \textbf{6.05} & 21.39          & 7.24          \\ \hline
Merge attention     & PPL         & 6.40          & 4.87          & 2.86          & 1.13          & 5.93          & 22.71          & 7.32          \\ \hline
Separate attention & task vector & 4.00            & 7.32          & \textbf{2.98} & 2.5           & 5.37          & 20.11          & 7.05          \\ \hline
Separate attention & PPL         & \textbf{6.80} & \textbf{7.92} & 2.88          & \textbf{2.95} & 4.74          & \textbf{23.21} & \textbf{8.08} \\ \hline
\end{tabular}
}
\caption{\label{table:moe_wo_finetune} \textbf{Performance of proposed merging and routing methods for MoE without substantial fine-tuning and other baselines across six datasets.} Separating attention layers and perplexity routing heuristics get the best average performance.}
\end{table}


\noindent \textbf{Proposed MoE method without fine-tuning outperforms the dense merging baseline.} \quad From Table \ref{table:moe_wo_finetune}, we observe that using the PPL routing heuristic and separating attention layers achieves the best average results among all baseline methods. Compared to Random Routing and the SoTA dense merging method (Dare), our best method - PPL routing + separating attention layers - yields relative improvements of 16.8\% and 13.6\%, respectively. The superior performance of PPL routing aligns with Figure \ref{fig:routing_heuristic} in Appendix \ref{sec:supp_routing}, where PPL routing more accurately directs input to the appropriate experts. 
Moreover, the better results of separating attention layers support our expectation that this approach resolves the inconsistency of task vector counts, as discussed in Section \ref{sec:merging_wo_ft}.

\subsection{Heterogeneous Model Merging}


\textbf{MoE merged with heterogeneous models outperforms the corresponding experts.} \quad
After showing the superiority of our homogeneous model merging method, our next question is whether the proposed heterogeneous expert merging is also effective.
We present the performance of the dense, MoE and baseline methods in Table \ref{table:hetero_results}. The details of training cost for each method are presented in Table \ref{table:4_training_cost} in Appendix.

\begin{table}[!htb]
\centering
\resizebox{\columnwidth}{!}{%
\begin{tabular}{lccccccc}
\toprule
Method                                                           & MBPP           & HumanEval      & MATH          & GSM8K         & NQ            & TriviaQA      & Avg.           \\ \hline
\multicolumn{8}{c}{\textbf{Dense Model}}                                                                                                                                            \\ \hline
\llamab                                                      & 4.60           & 3.04           & 2.42          & 1.44          & 6.61          & 26.72         & 7.47           \\ \hline
Base TinyLlama                                                   & 5.40           & 5.27           & 2.26          & 2.2           & 8.53          & 34.27         & 9.66           \\ \hline
Base Olmo                                                        & 2.80           & 2.64           & 2.46          & 2.42          & 6.16          & 29.21         & 7.62           \\ \hline
Code \llama                                                       & 10.20          & 8.53           & 2.42          & 2.57          & 3.11          & 16.7          & 7.26           \\ \hline
Math TinyLlama                                                   & 15.60          & 9.76           & 4.18          & 5.91          & 6.05          & 21.12         & 10.44          \\ \hline
Math Olmo                                                        & 0.00           & 0.00           & 4.82          & 5.08          & 3.61          & 11.25         & 4.13           \\ \hline
Knowledge \llama                                                  & 3.60            & 4.26           & 2.62          & 2.04          & 5.65          & 28.71         & 7.81           \\ \hline

\multicolumn{8}{c}{\textbf{Homogeneous Expert Merging}}                                                                                                                                              \\ \hline
\begin{tabular}[c]{@{}l@{}}3-expert MoE \\ (same data)\end{tabular}  & 9.14           & 10.8           & 4.42          & 5.16          & 6.95          & 26.78         & 10.54          \\ \hline
\begin{tabular}[c]{@{}l@{}}3-expert MoE \\ (w/o math)\end{tabular}   & 12.00             & 9.76           & 2.38          & 1.74          & 6.22          & \textbf{33.20} & 10.88          \\ \hline

\multicolumn{8}{c}{\textbf{Heterogeneous Expert merging}}                                                                                                                                              \\ \hline
\begin{tabular}[c]{@{}l@{}}(Ours) MoE w/ \\ Math Olmo\end{tabular}      & 13.60           & 10.98          & 4.86          & 6.14          & 5.43 & 26.01         & 11.17          \\ \hline

\begin{tabular}[c]{@{}l@{}} (Ours) MoE w/ \\ Math TinyLlama\end{tabular} & \textbf{15.80} & \textbf{11.59} & \textbf{5.42} & \textbf{6.29} & \textbf{8.25} & 32.71         & \textbf{13.34} \\ \bottomrule
\end{tabular}
}
\caption{\label{table:hetero_results} \textbf{Performance of proposed heterogeneous merged MoE and other baselines.} The merged MoE is comparable or outperform the dense or 3-expert baselines on the benchmark from the corresponding domain.}
\end{table}


Table \ref{table:hetero_results} shows that our merged MoE models are comparable to or outperform the domain expert models in their respective domains. For instance, the MoE merged with Math Olmo and Math TinyLlama achieves 6.14\% and 6.29\% accuracy on GSM8K, compared to 5.91\% and 5.08\% for their dense counterparts. On average, our MoEs with Olmo and TinyLlama improves performance by 43.02\% and 27.78\% relative to the best dense experts, respectively. Both MoEs with heterogeneous experts also outperform the 3-expert MoE baseline, particularly in math, highlighting the effectiveness of including math experts in the pipeline.

\begin{figure}[!t]
    \centering
    \begin{subfigure}[b]{0.48\columnwidth}
        \centering
        \includegraphics[width=\textwidth]{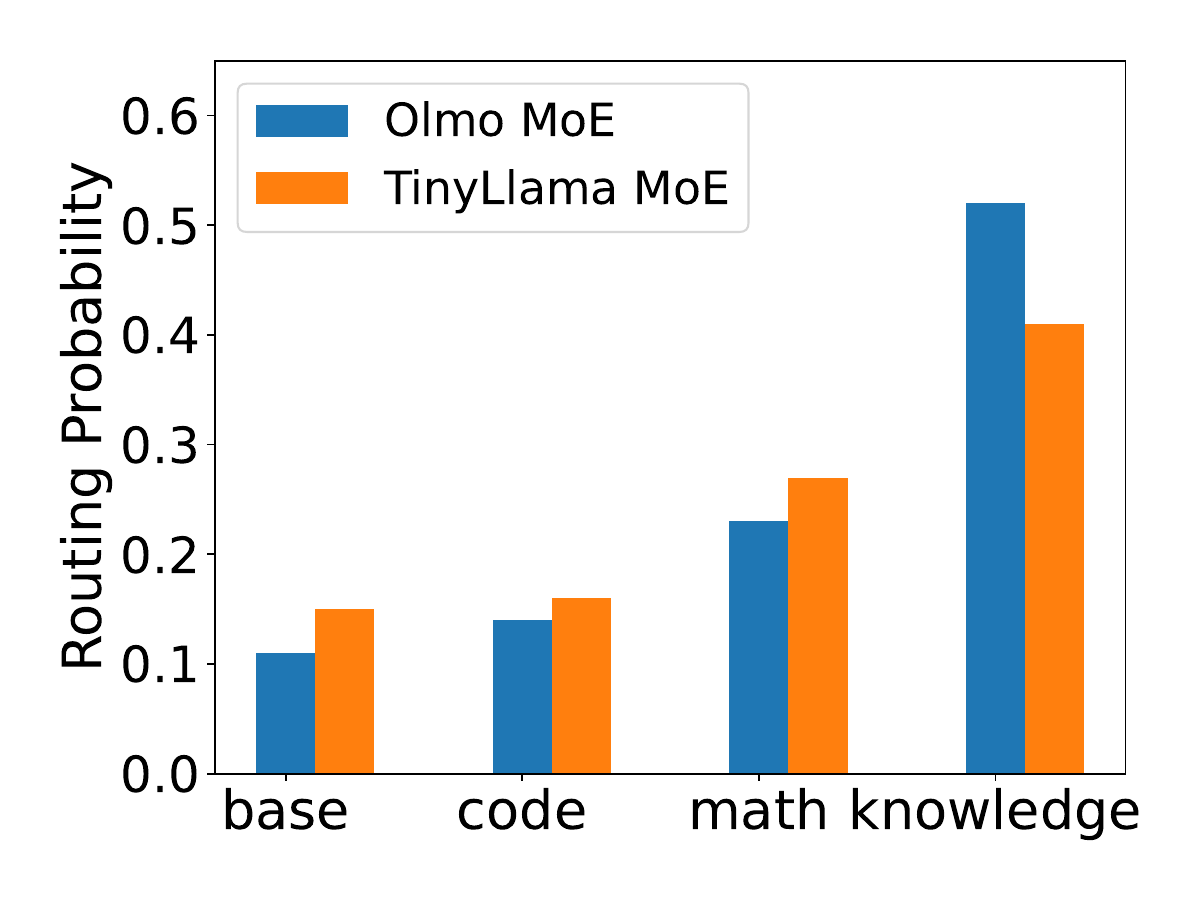}
        \caption{GSM8K}
        \label{fig:hetero_gsm8k_route}
    \end{subfigure}
    \hfill
    \begin{subfigure}[b]{0.48\columnwidth}
        \centering
        \includegraphics[width=\textwidth]{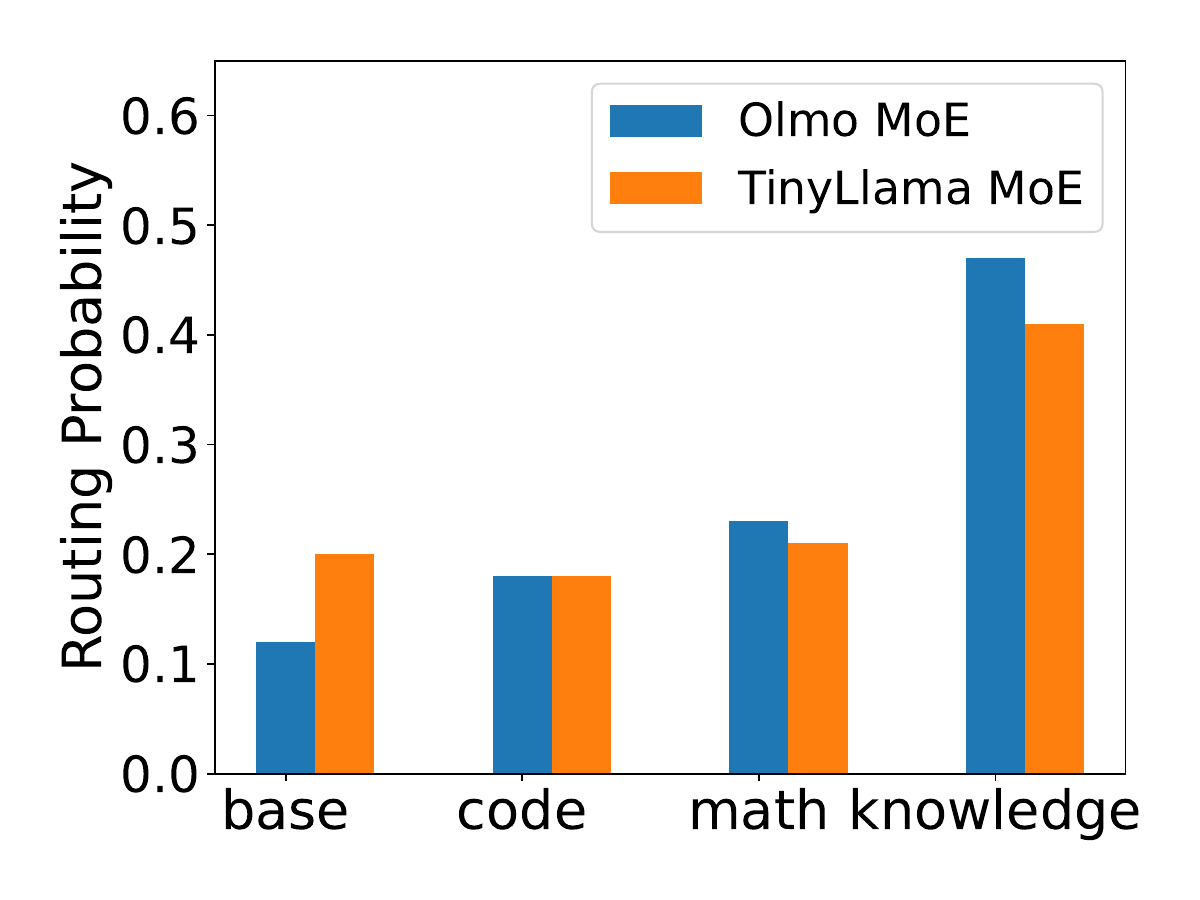}
        \caption{MATH}
        \label{fig:hetero_math_route}
    \end{subfigure}
    \caption{Routing probability of experts on GSM8K and MATH for the MoE w/ Olmo and MoE w/ TinyLlama.}
    \label{fig:hetero_routing_prob}
\end{figure}

\noindent \textbf{MoE merged with heterogeneous experts show the desired routing patterns in most cases.} \quad We also perform a similar routing analysis as described in Section \ref{sec:rq1}. We visualize the routing probability of two MoEs when evaluating on GSM8K and MATH datasets in Figure \ref{fig:hetero_routing_prob} and for other datasets, we visualize the results in Figure \ref{fig:supp_hetero_routing_prob} in Appendix \ref{sec:supp_routing}.

As shown in Figures \ref{fig:hetero_routing_prob} and \ref{fig:supp_hetero_routing_prob}, most tokens in the coding and knowledge datasets are routed to the corresponding experts. However, unlike homogeneous model merging where the math expert has the highest routing probability for math datasets, Math Olmo or Math TinyLlama ranks second. This discrepancy is likely due to the difference in embedding outputs between the MoE and expert models. Since the MoE's embedding layer is merged from 3 \llama models and 1 other model, its output is closer to that of the \llama models, making the router more likely to select them. Adding a load balancing loss is a possible solution to address this issue \cite{sukhbaatar2024branchtrainmixmixingexpertllms, fedus2022switch}, ensuring a more uniform routing distribution. We leave this for future exploration
\section{Conclusion}


In this paper, we propose novel methods to address challenges in the current MoE merging literature. For homogeneous experts, we replace average merging in non-FFN layers with more advanced methods to reduce parameter interference. We also explore merging models into an MoE without post-merge fine-tuning. For heterogeneous experts, we introduce a method using projectors and sequence-level routing networks to combine models with different architectures. Extensive empirical evaluations show that our approach significantly improves MoE performance across multiple datasets.

\section{Limitation}

One of the limitations of the proposed merging methods with heterogeneous experts is that the merged MoE model has more parameters when the BTX merging, since we do not merge the attention layers. For example, for our 4 $\times$ 1B \llama MoE, the total parameter number is about 3.7 billion due to the non-FFNs layer merging but the total parameter number of the MoE after the heterogeneous merging method is near 4 billion. More parameters represent more costly fine-tuning and inference.

For our homogeneous merging method, we replace simple averaging with a more advanced merging method: Dare and Ties and fine-tune MoE models. There are still other merging methods, such as fisher merging \cite{matena2022merging} or Regmean \cite{jin2022dataless} methods. However, in the Ties and Dare paper \cite{yadav2024ties, yu2024language}, they have demonstrated the superiority of proposed merging methods over Regmean and finisher merging, so we leave the exploration of other merging methods to future work.

Moreover, using routing heuristics to process the input sequence introduces additional inference costs, as we first need to use the expert model to calculate the perplexity (PPL) or gradient. However, our routing heuristic requires only one additional forward pass, and considering the multiple forward passes during inference (forward pass number = the generate token number), the computational overhead for our method to enhance MoE performance without fine-tuning is minimal.

For all MoE fine-tuning, we utilize only the cross-entropy loss to do the auto-regression on the training data. Previous works showed that the load-balancing loss \cite{fedus2022switch, sukhbaatar2024branchtrainmixmixingexpertllms} may be beneficial to resolve the ``dead'' experts. From our routing analysis for the merged MoEs, we observe that merging with homogeneous experts gets the desirable patterns, where most tokens in one specific domain are gated to the corresponding expert. However, for heterogeneous experts, due to the different architecture and tokenizer of the math expert, the math expert does not get the highest routing probability in evaluating on GSM8K and MATH datasets. For the next step, we may need to add the load balancing loss for the fine-tuning of MoE with heterogeneous experts to develop more robust models \cite{zhou2024explore} and observe whether the routing patterns are more efficient.

Due to limitations of computation resources, we only experimented with three domains and 1b LLMs. Incorporating larger models and more domains, such as legal, medical, or multilingual, can benefit future studies. Furthermore, our method can be extended to multimodal MoE by incorporating vision audio or graph experts \cite{wang2024mementos, wang2024enhancing, li2024uni, zhu2024multimodal}.

In addition to directly merging models with different architectures with additional projectors, there is another direction to first distill the knowledge of experts to student models with the same architecture \cite{wan2024knowledge, zhou2024teaching, li2025benchmarkevaluationsapplicationschallenges, zhou2023scalable, zhou2024multi} and merge student models together to an MoE. We leave the exploration of this direction to future work. 

\section*{Acknowledgments}
We would like to thank the anonymous reviewers as well
as Saleh Soltan, He Xie, Venkatesh Elango, Wael Hamza, Paiheng Xu, and Xiyao Wang for providing helpful comments and suggestions. 
\bibliographystyle{acl_natbib}
\bibliography{custom}

\appendix
\section{Implementation Details}
\label{sec:implement}
For our Base-1B models, we utilize the Llama-2 architecture \cite{wu2024llama} with layer number 24 and hidden dimension 2048. The open-source TinyLlama-1.1B model contains 22 layers and the hidden dimension is 2048. For the open-source Olmo-1B model, it has 16 layers and the hiddn dimension is 2048.

In our experiments, we use top-2 routing for MoE models. For Dare-merging and Ties merging (both dense and MoE), we set the scaling term $\lambda$ to $\frac{1}{3}$ and the retain ratio $p$ of the model parameters of two methods are set to $80\%$ to gain the optimal performance, according to our preliminary exploration. For inference, we set the temperature to 0.0 for greedy decoding, and the maximal number of generated tokens is 512. For CPT and fine-tuning of MoE and dense models, we set the learning rate to 1e-5 and the weight decay is 0.01.

\section{Data mixture}
\label{sec:data_mix}

In Table \ref{table:data_mix}, we present the data ratios to CPT or fine-tune the dense or MoE models.
For fine-tuning the MoE model, we sample datasets that are used to train all experts and the base model with the same probabilities as described in \citet{sukhbaatar2024branchtrainmixmixingexpertllms}.

\begin{table}[!htb]
\centering
\resizebox{\columnwidth}{!}{%
\begin{tabular}{lccccc}
\toprule
                & Base    & Math    & Code    & Knowledge & Finetune MoE \\ \midrule
Wiki1           & 0.85\%  & 0.17\%  & 0.17\%  & 8.00\%    & 1.11\%       \\ \midrule
Wiki2           & 0.00\%  & 0.00\%  & 0.00\%  & 8.00\%    & 0.82\%       \\ \midrule
Arxiv           & 9.37\%  & 1.87\%  & 1.87\%  & 7.94\%    & 3.94\%       \\ \midrule
CommonCrawl     & 27.92\% & 5.58\%  & 5.58\%  & 23.65\%   & 11.74\%      \\ \midrule
C4              & 54.60\% & 10.93\% & 10.93\% & 46.26\%   & 22.97\%      \\ \midrule
StackExchange   & 7.26\%  & 1.45\%  & 1.45\%  & 6.15\%    & 3.05\%       \\ \midrule
Open   Web Math & 0.00\%  & 80.00\% & 0.00\%  & 0.00\%    & 24.13\%      \\ \midrule
GitHub          & 0.00\%  & 0.00\%  & 80.00\% & 0.00\%    & 32.25\% \\
\bottomrule
\end{tabular}
}
\caption{\label{table:data_mix} Data source and weights for CPT or fine-tune MoE or dense models. Wiki1 represents the first half of Wikipedia data for pretraining the base model and Wiki2 represents the second half of Wikipedia data for CPT the knowledge expert.}
\end{table}

\section{Task Vector Routing Heuristic}
\label{sec:task_vector_routing}

Our second approach is to identify the input domain and assign the input to experts trained in that domain. The core idea is that an expert's task vector, defined as the difference between its parameters and the base model, represents the cumulative gradient of the base model on the expert's training data. For a given input, we first compute the base model's gradient on that input and compare it to the task vectors of each expert. A higher similarity between the gradient and a task vector suggests the input is closer to the expert's training data.

With the task vectors $\tau_1, \tau_2, \dots, \tau_l$ for $l$ experts and inference input $x_{inf}$, the loss function $\mathcal{L}$ and the base model parameters $\theta_b$, we first compute the gradient ($g_{inf}$) of the loss function with respect to the base model parameters as: $g_{inf} = \nabla_{\theta_b}\mathcal{L}(x_{inf})$.

The routing heuristic decides the experts and weights with the cosine similarity (Sim) as below:
\begin{equation*}
    \resizebox{\columnwidth}{!}{%
        $\alpha = \text{SoftMax}(\text{top-K}(\text{Sim}(g_{inf}, \tau_1), \dots, \text{Sim}(g_{inf}, \tau_l)))$%
        }
\end{equation*}




\section{Supplementary Results}
\label{sec:supp_routing}

In this section, we present the supplementary analysis of the routing probability for each research question.

\begin{figure}[!t]
    \centering
    \begin{subfigure}[b]{0.48\columnwidth}
        \centering
        \includegraphics[width=\textwidth]{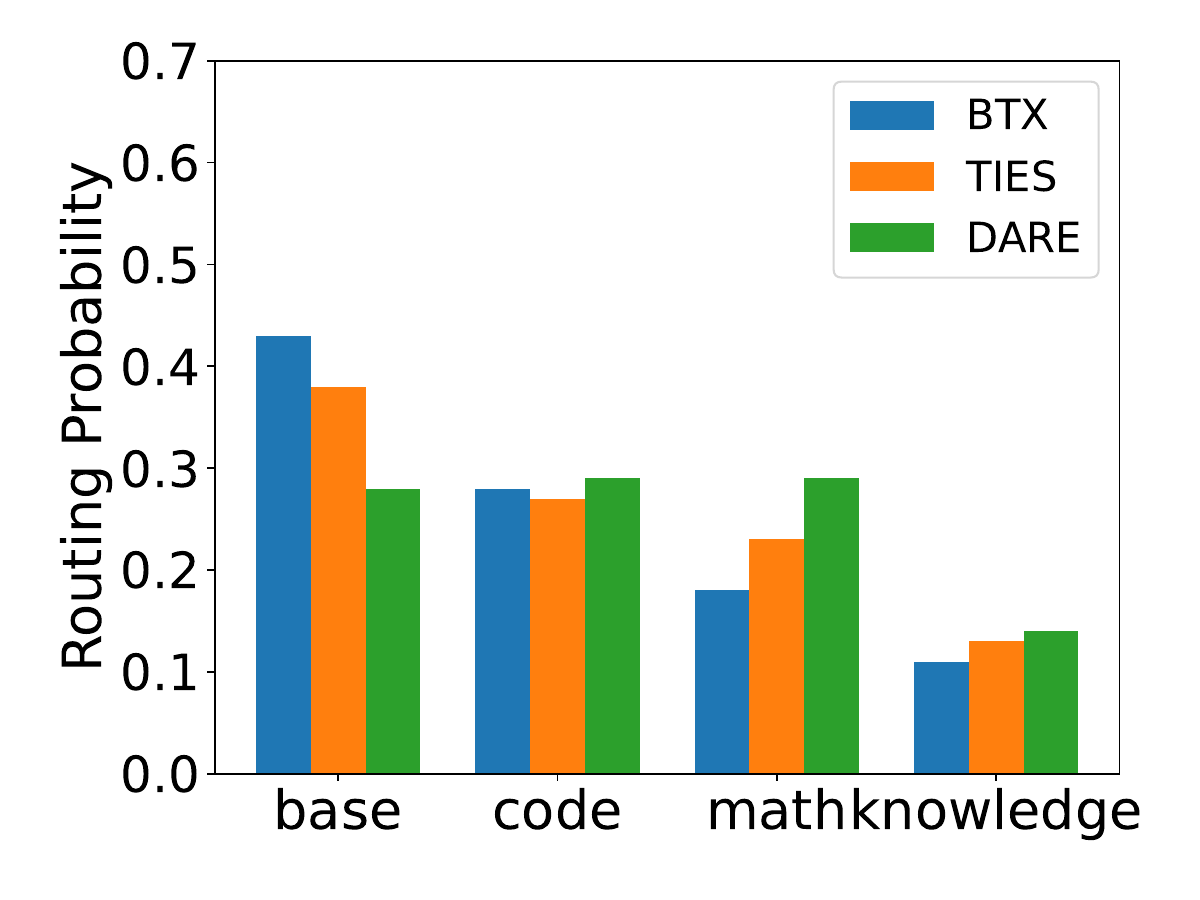}
        \caption{MBPP}
    \end{subfigure}
    \hfill
    \begin{subfigure}[b]{0.48\columnwidth}
        \centering
        \includegraphics[width=\textwidth]{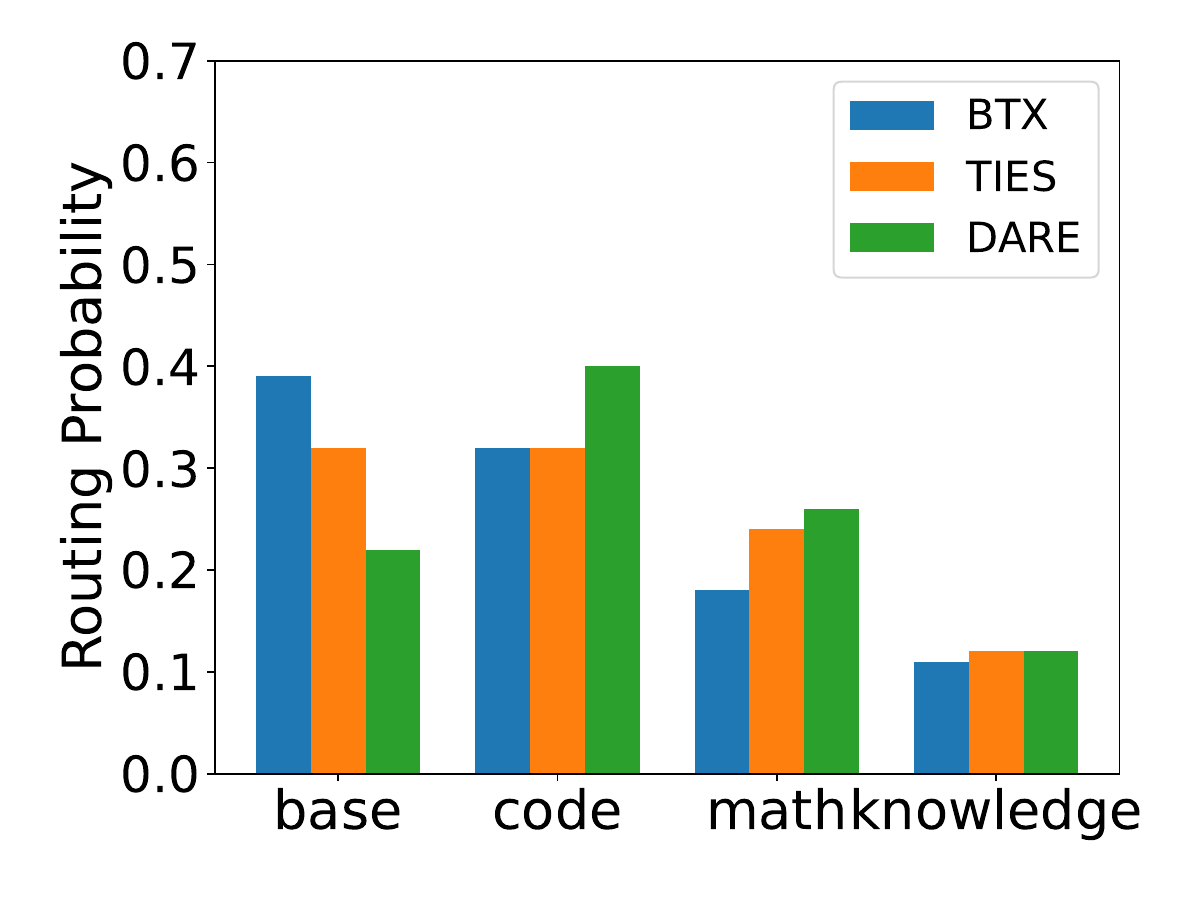}
        \caption{HumanEval}
    \end{subfigure}
    \hfill
    \begin{subfigure}[b]{0.48\columnwidth}
        \centering
        \includegraphics[width=\textwidth]{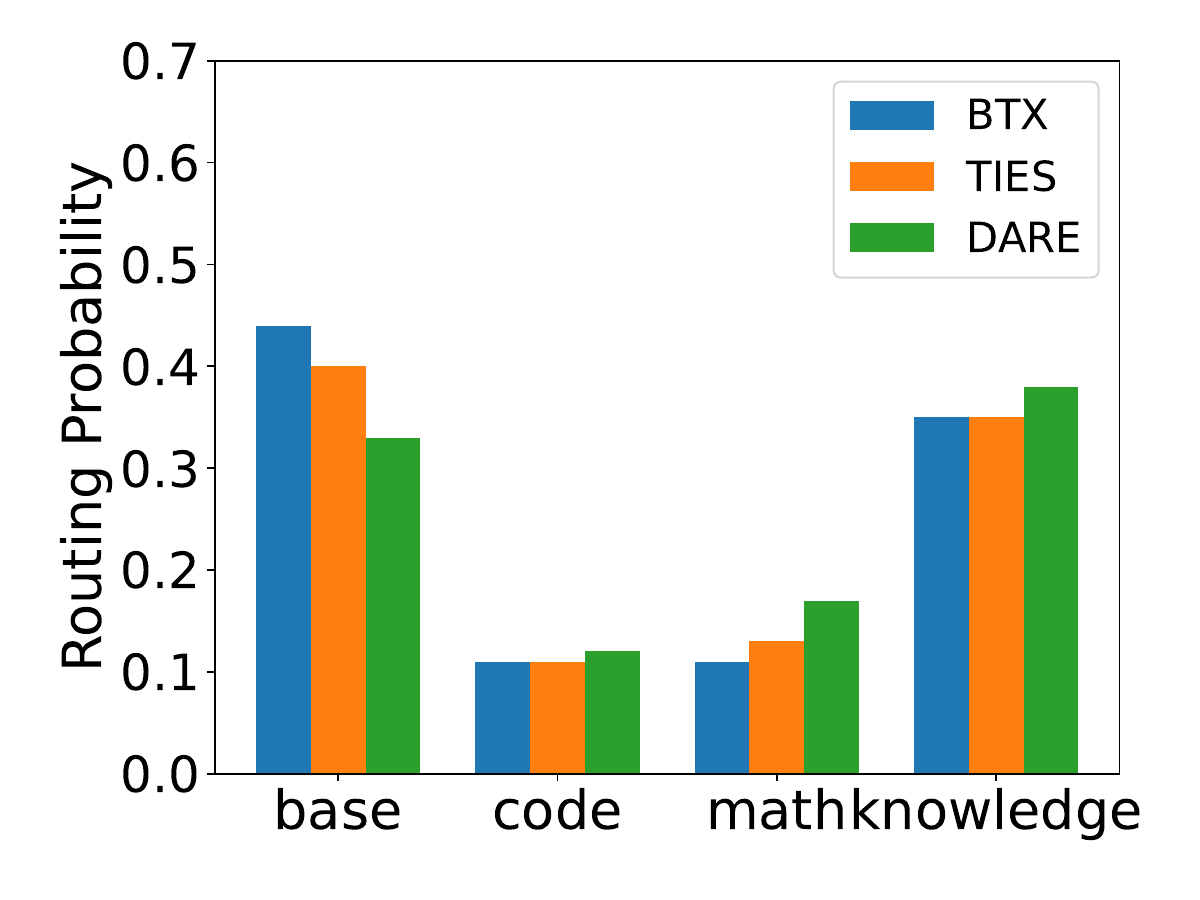}
        \caption{Natural Questions}
    \end{subfigure}
    \hfill
    \begin{subfigure}[b]{0.48\columnwidth}
        \centering
        \includegraphics[width=\textwidth]{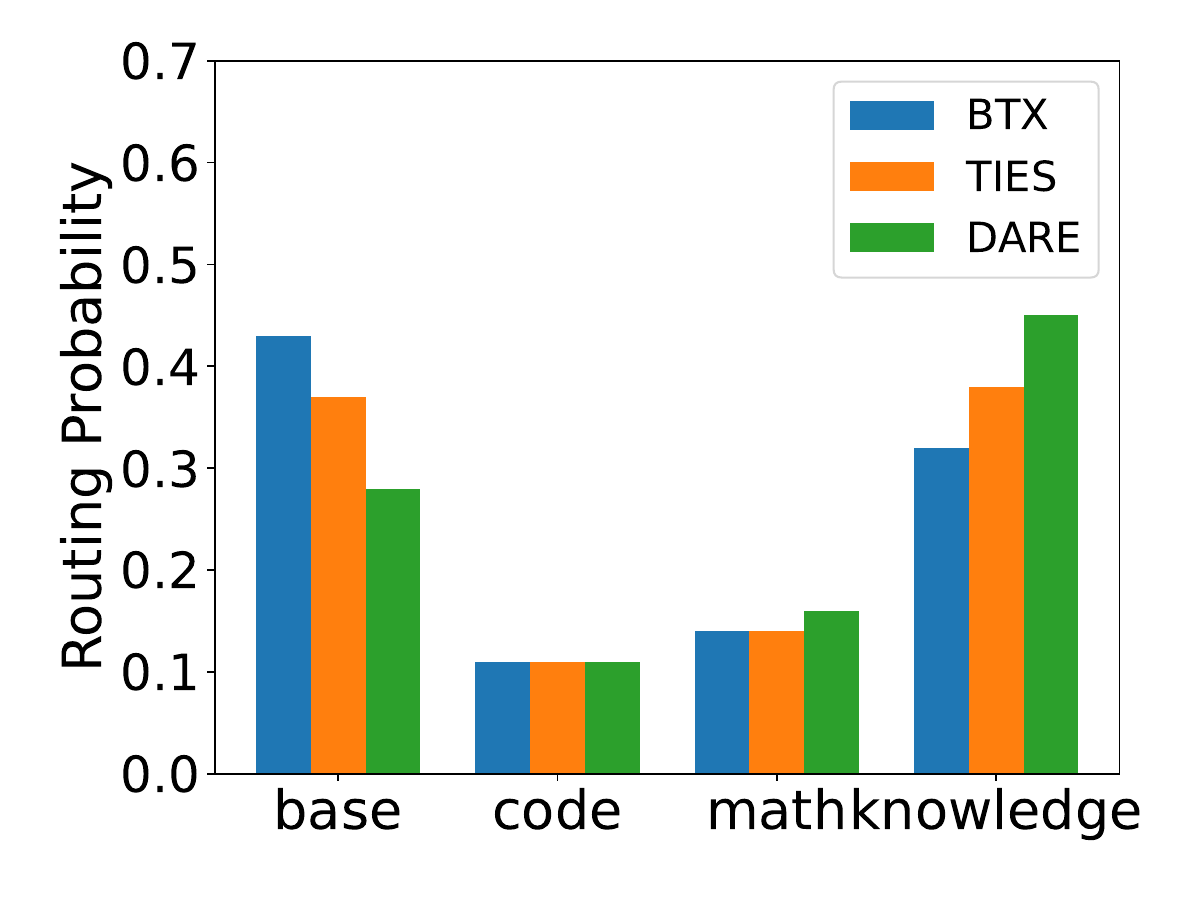}
        \caption{TriviaQA}
    \end{subfigure}
    \caption{Routing probability of experts on MBPP, HumanEval, Natural Questions and TriviaQA for different merging methods.}
    \label{fig:supp_routing_prob}
\end{figure}

\begin{figure}[!t]
    \centering
    \begin{subfigure}[b]{0.48\columnwidth}
        \centering
        \includegraphics[width=\textwidth]{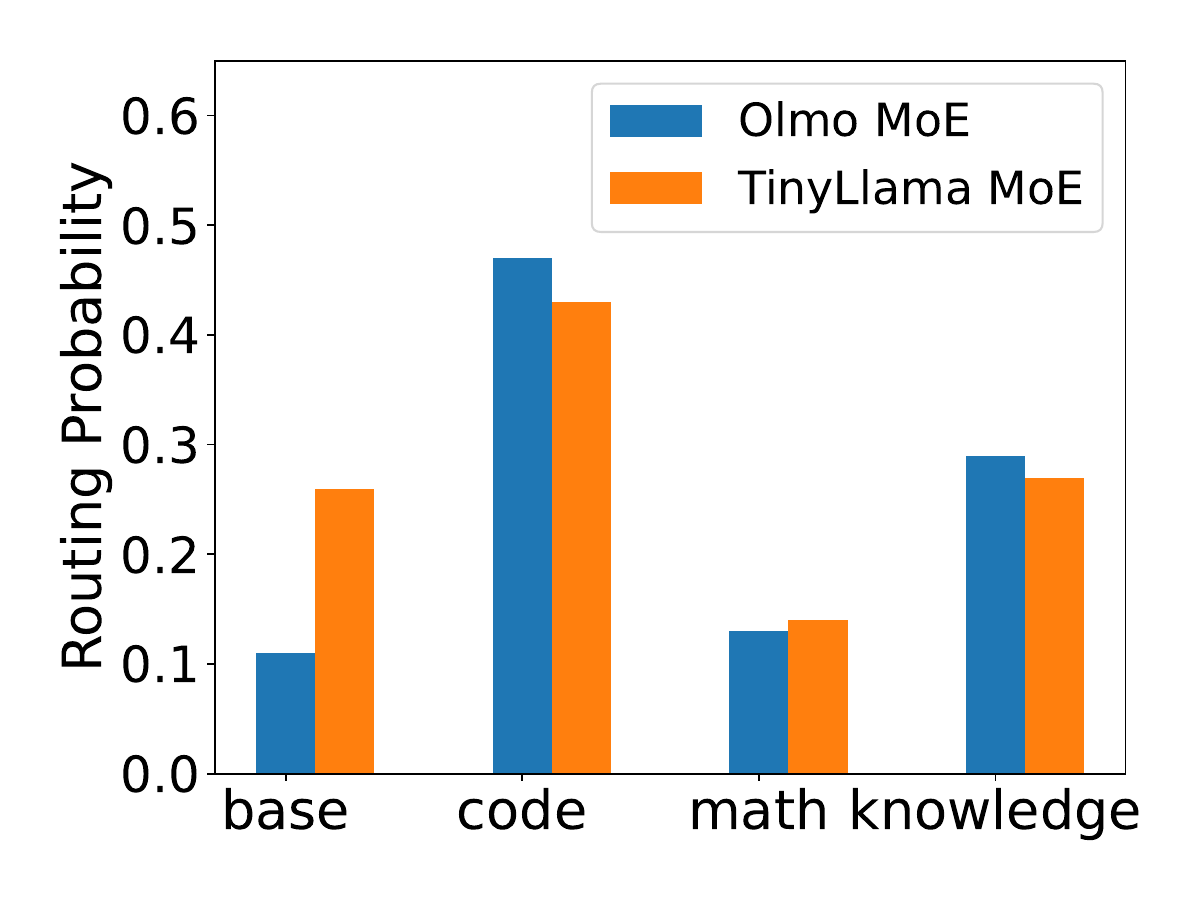}
        \caption{MBPP}
    \end{subfigure}
    \begin{subfigure}[b]{0.48\columnwidth}
        \centering
        \includegraphics[width=\textwidth]{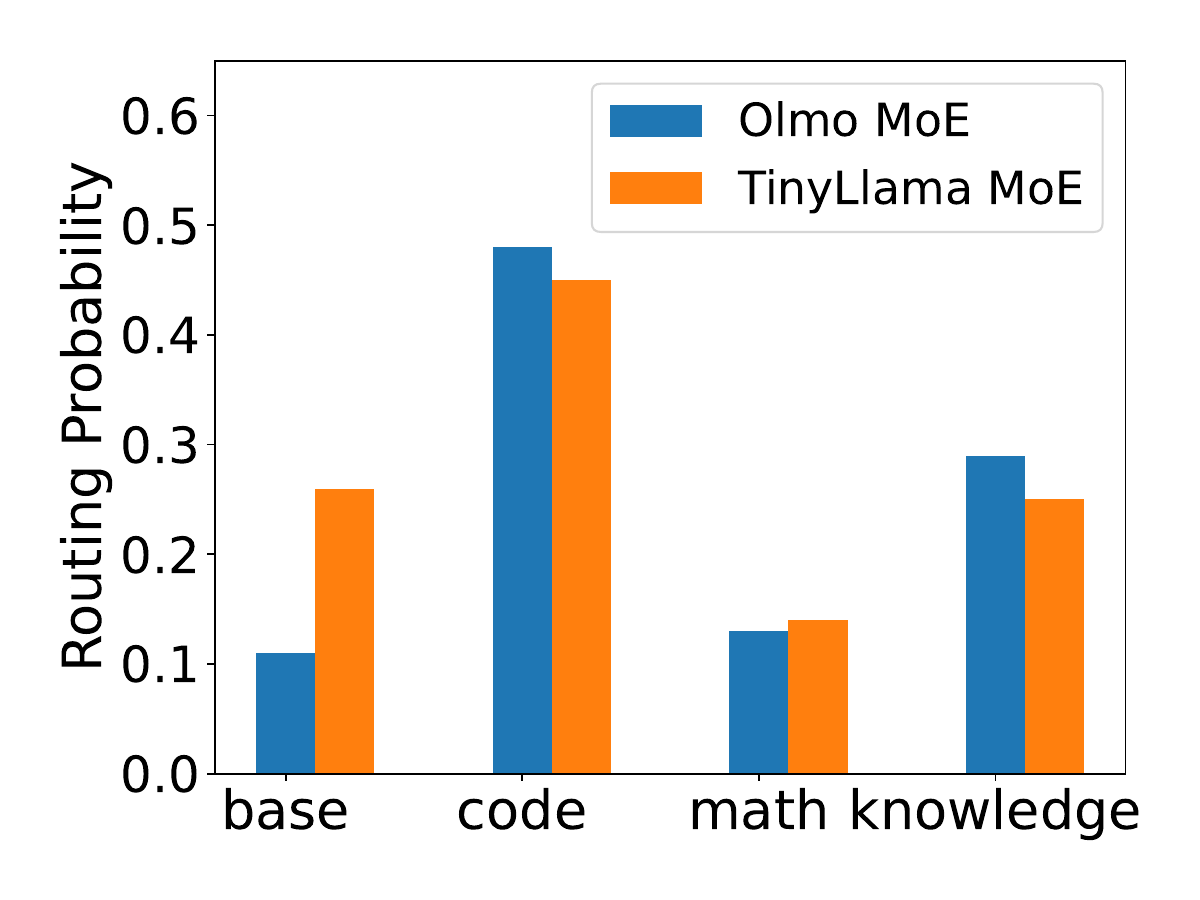}
        \caption{HumanEval}
    \end{subfigure}
    \hfill
    \begin{subfigure}[b]{0.48\columnwidth}
        \centering
        \includegraphics[width=\textwidth]{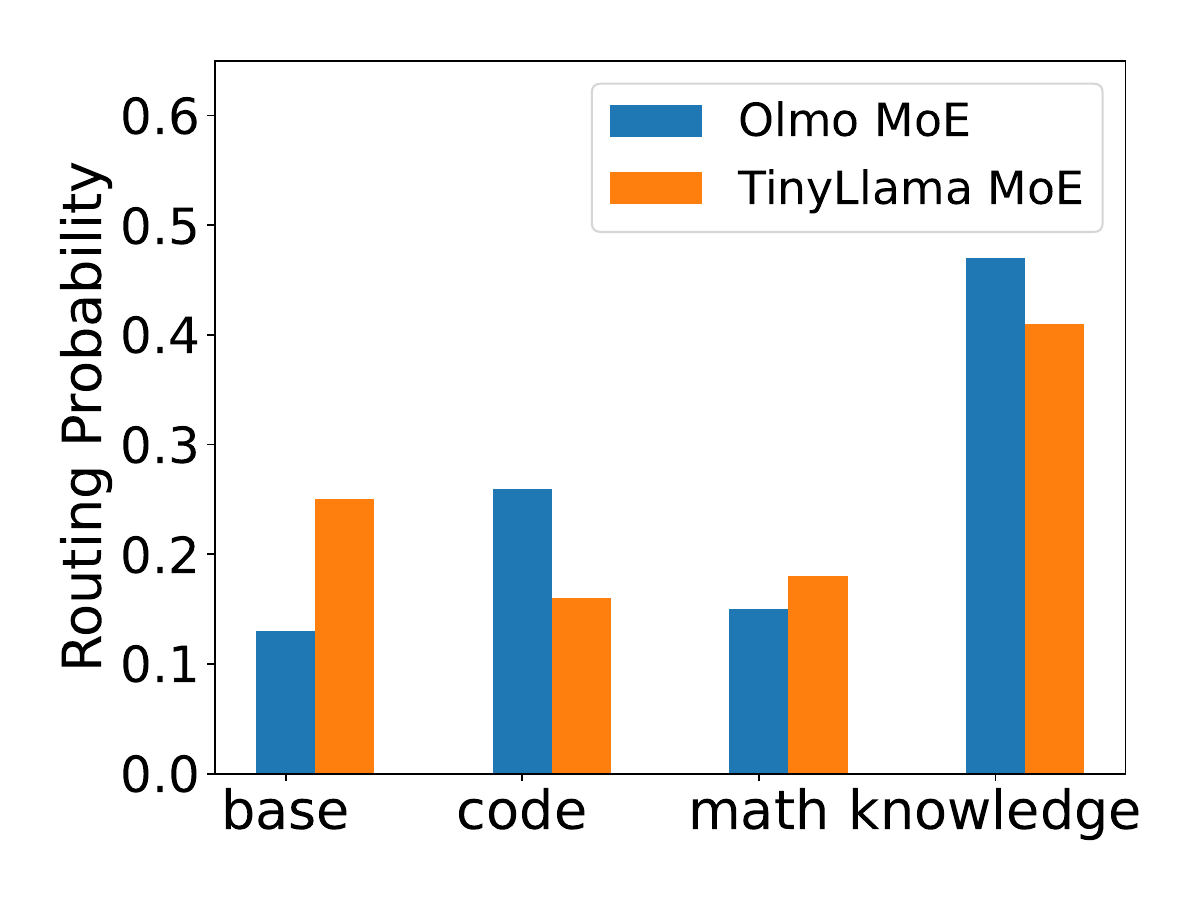}
        \caption{Natural Questions}
    \end{subfigure}
    \begin{subfigure}[b]{0.48\columnwidth}
        \centering
        \includegraphics[width=\textwidth]{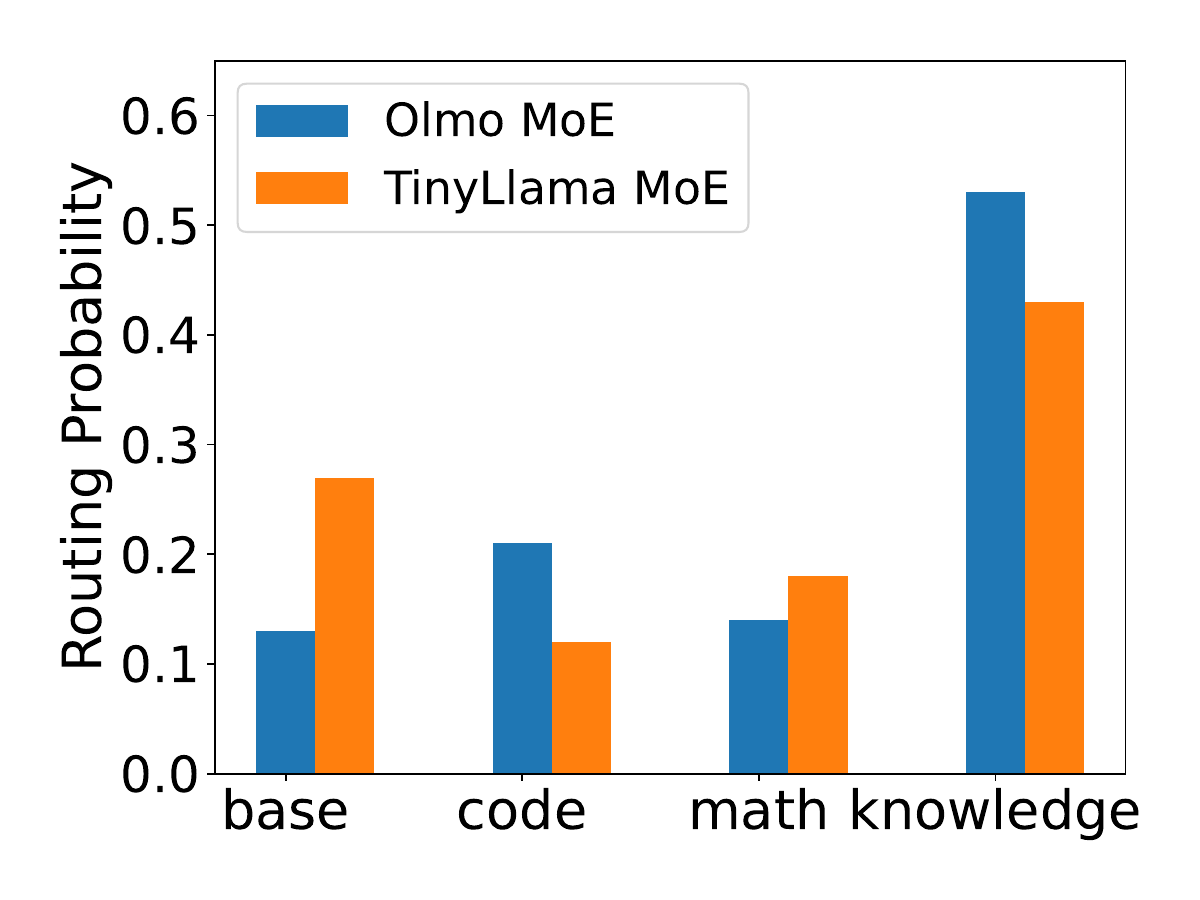}
        \caption{TriviaQA}
    \end{subfigure}
    \caption{Routing probability of experts on MBPP, HuamnEval, Natural Questions and TriviaQA for the MoE w/ Olmo and MoE w/ TinyLlama.}
    \label{fig:supp_hetero_routing_prob}
\end{figure}

\begin{figure}[!t]
    \centering
    \begin{subfigure}[b]{0.48\columnwidth}
        \centering
        \includegraphics[width=\textwidth]{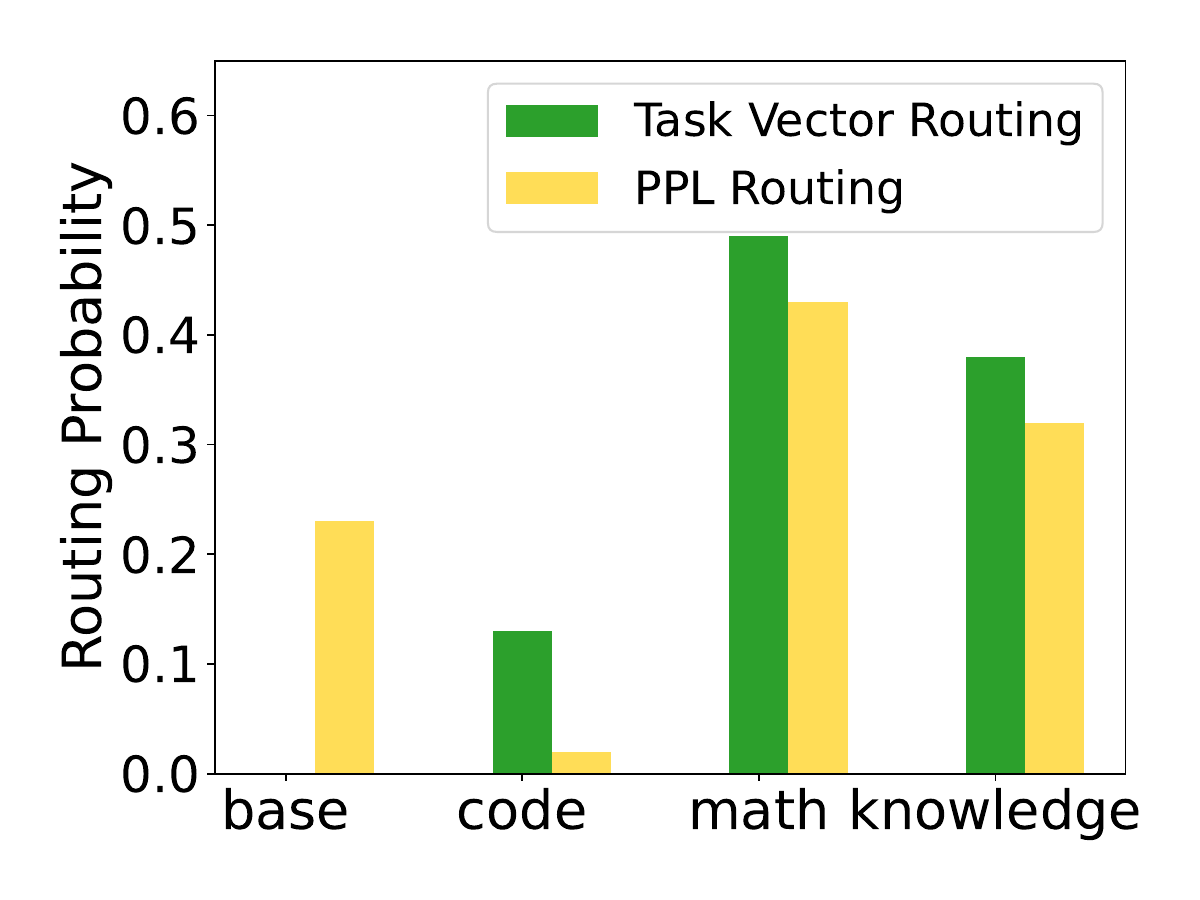}
        \caption{GSM8K}
    \end{subfigure}
    \begin{subfigure}[b]{0.48\columnwidth}
        \centering
        \includegraphics[width=\textwidth]{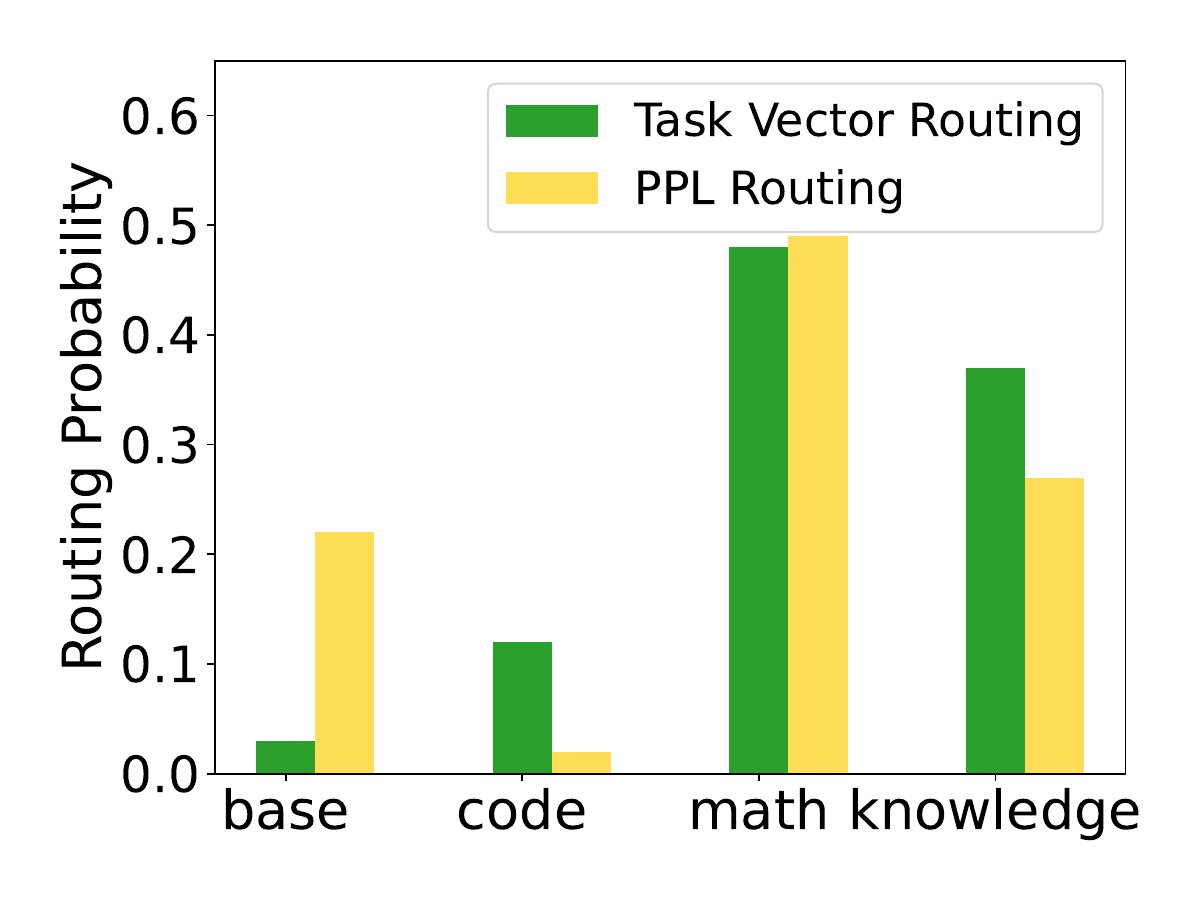}
        \caption{MATH}
    \end{subfigure}
    \hfill
    \begin{subfigure}[b]{0.48\columnwidth}
        \centering
        \includegraphics[width=\textwidth]{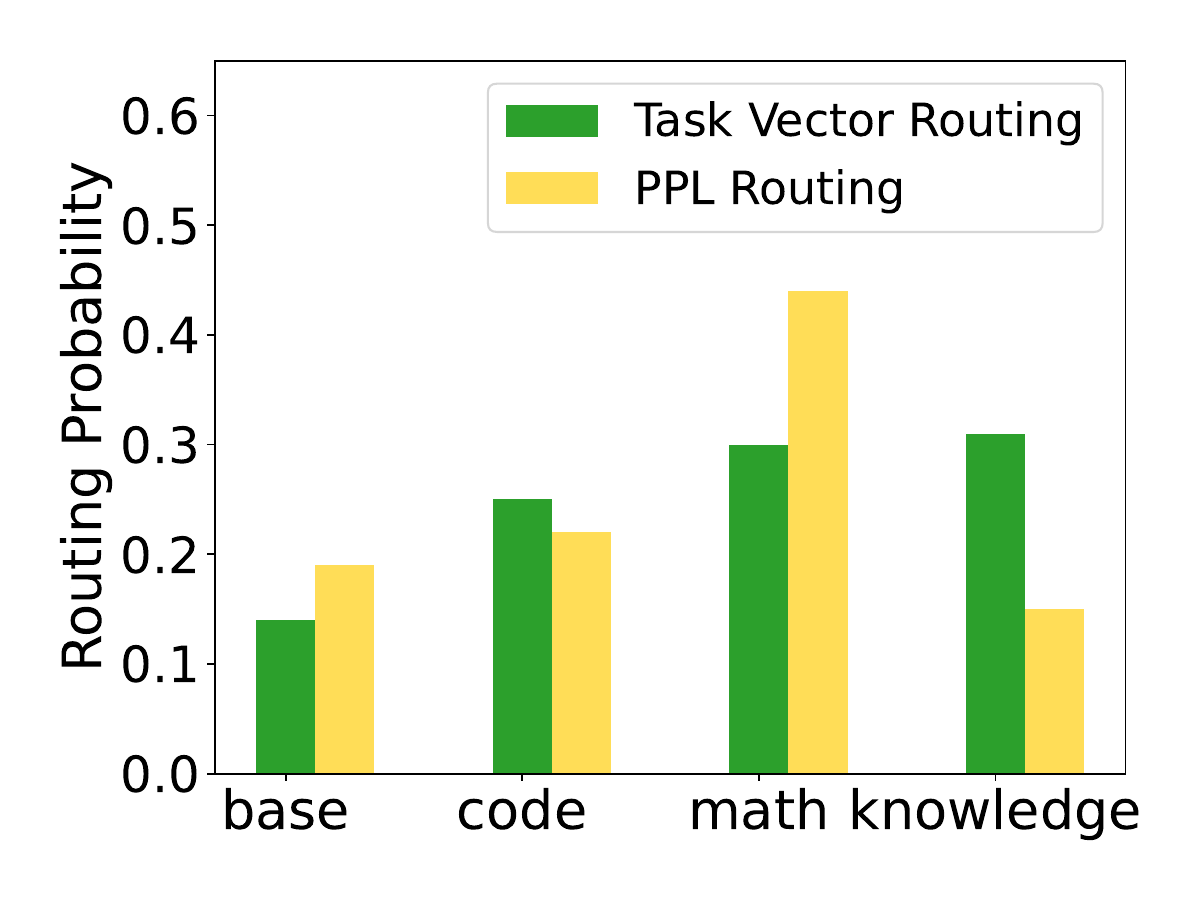}
        \caption{MBPP}
    \end{subfigure}
    \begin{subfigure}[b]{0.48\columnwidth}
        \centering
        \includegraphics[width=\textwidth]{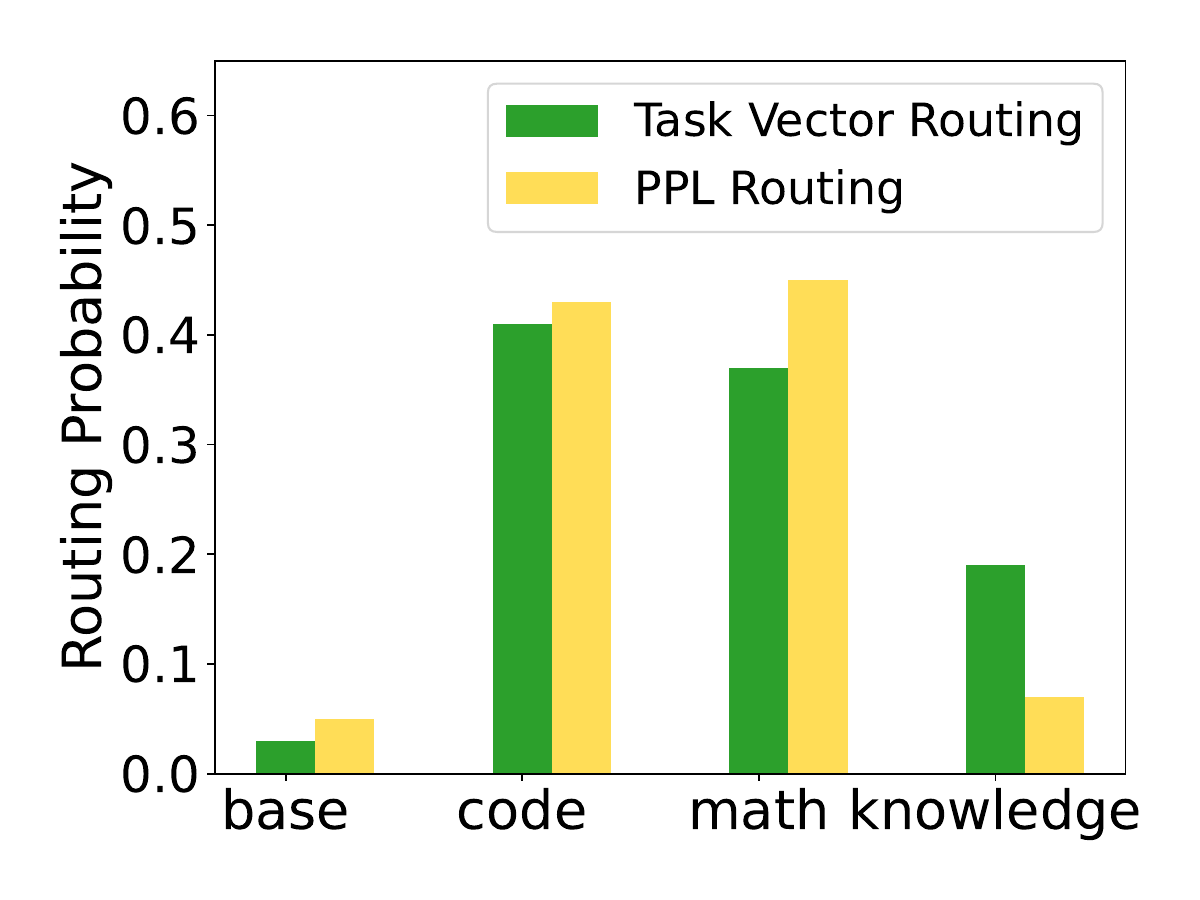}
        \caption{HumanEval}
    \end{subfigure}

    \centering
    \begin{subfigure}[b]{0.48\columnwidth}
        \centering
        \includegraphics[width=\textwidth]{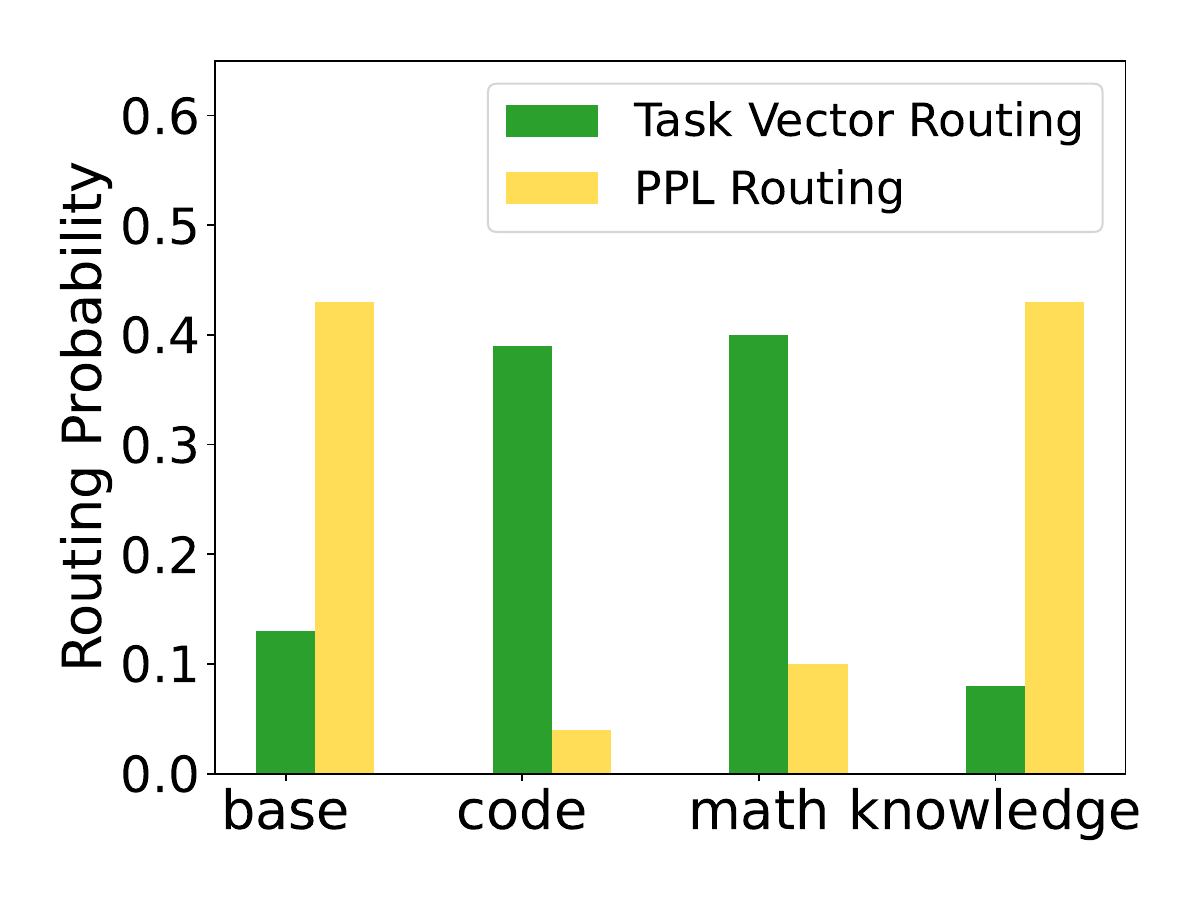}
        \caption{Natural Questions}
    \end{subfigure}
    \begin{subfigure}[b]{0.48\columnwidth}
        \centering
        \includegraphics[width=\textwidth]{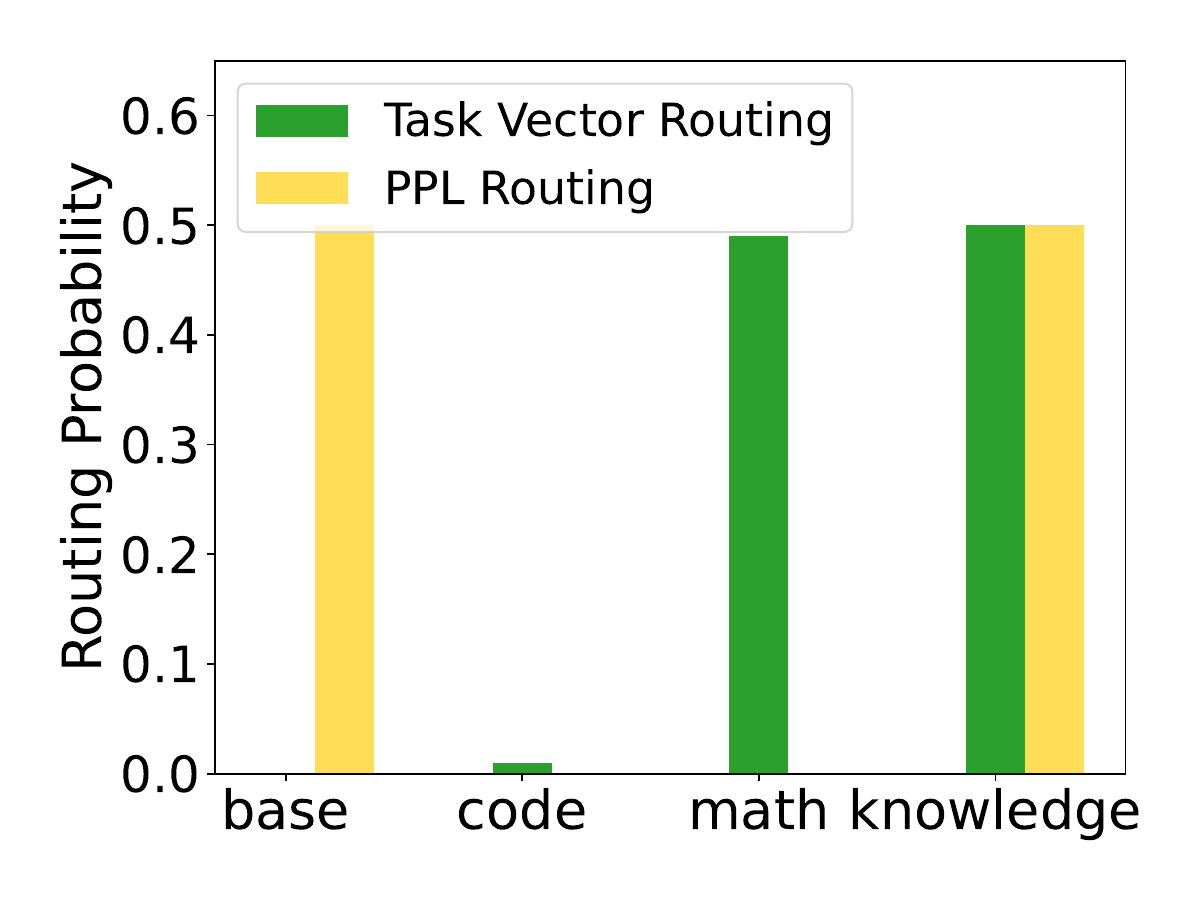}
        \caption{TriviaQA}
    \end{subfigure}
    \caption{\label{fig:routing_heuristic} Routing probability of tow routing heuristics for each dataset.}
\end{figure}

\begin{figure}[!t]
    \centering
    \begin{subfigure}[b]{0.48\columnwidth}
        \centering
        \includegraphics[width=\textwidth]{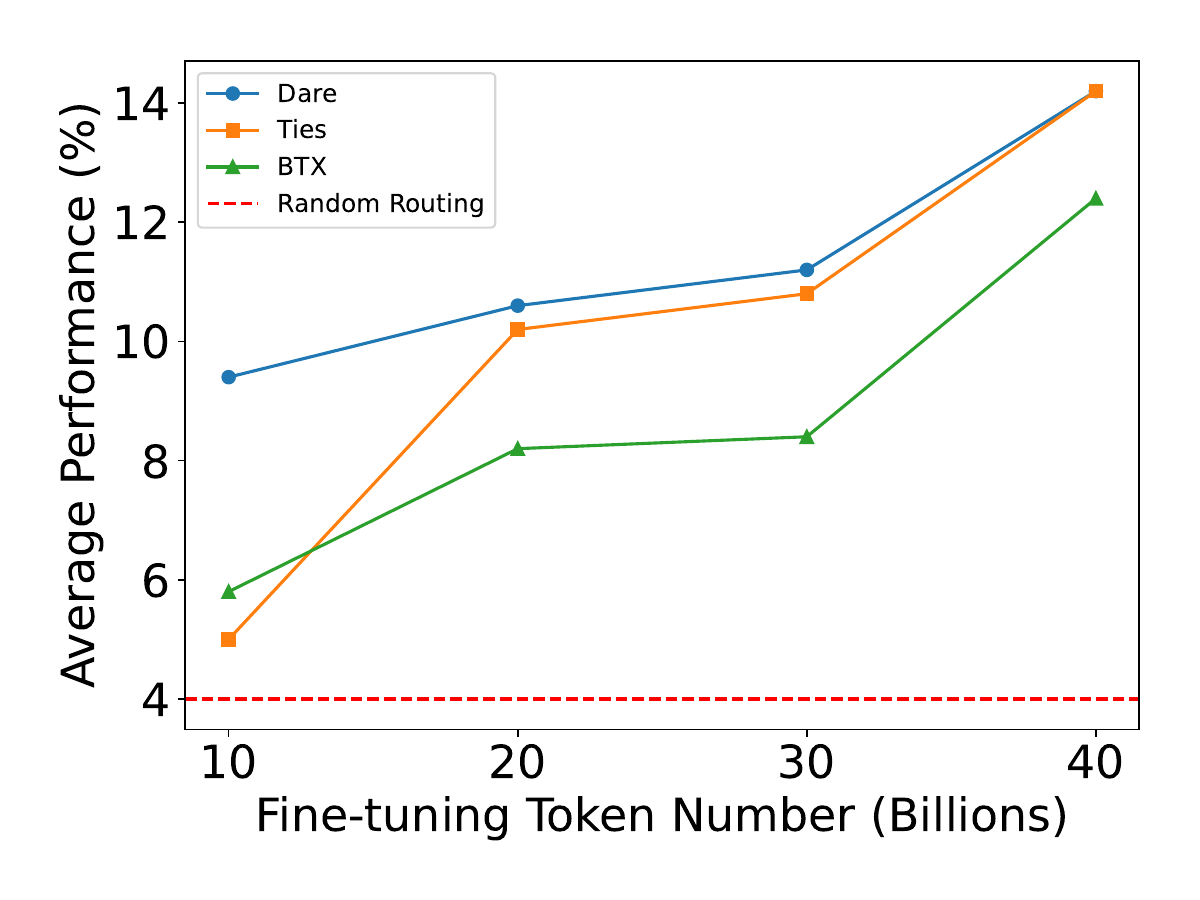}
        \caption{MBPP}
    \end{subfigure}
    \begin{subfigure}[b]{0.48\columnwidth}
        \centering
        \includegraphics[width=\textwidth]{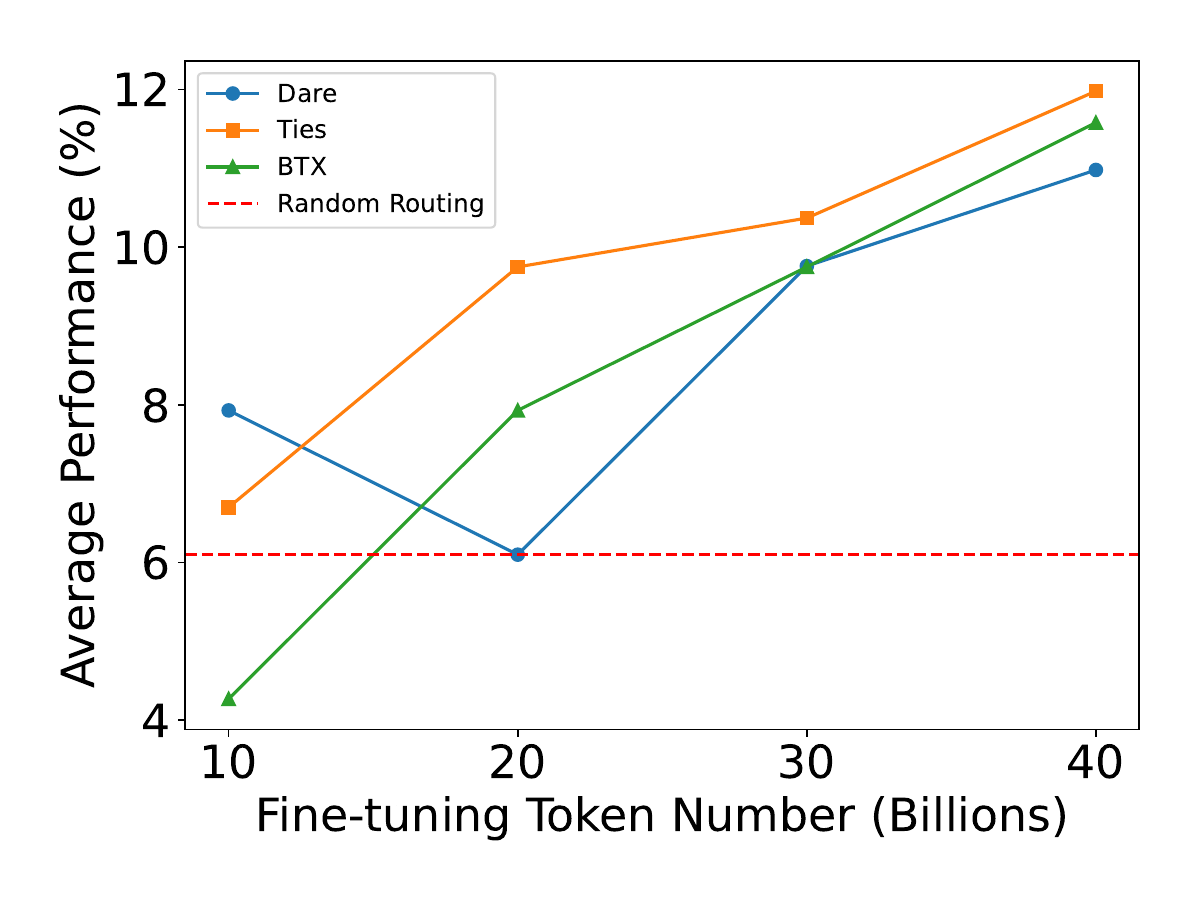}
        \caption{HumanEval}
    \end{subfigure}
    \begin{subfigure}[b]{0.48\columnwidth}
        \centering
        \includegraphics[width=\textwidth]{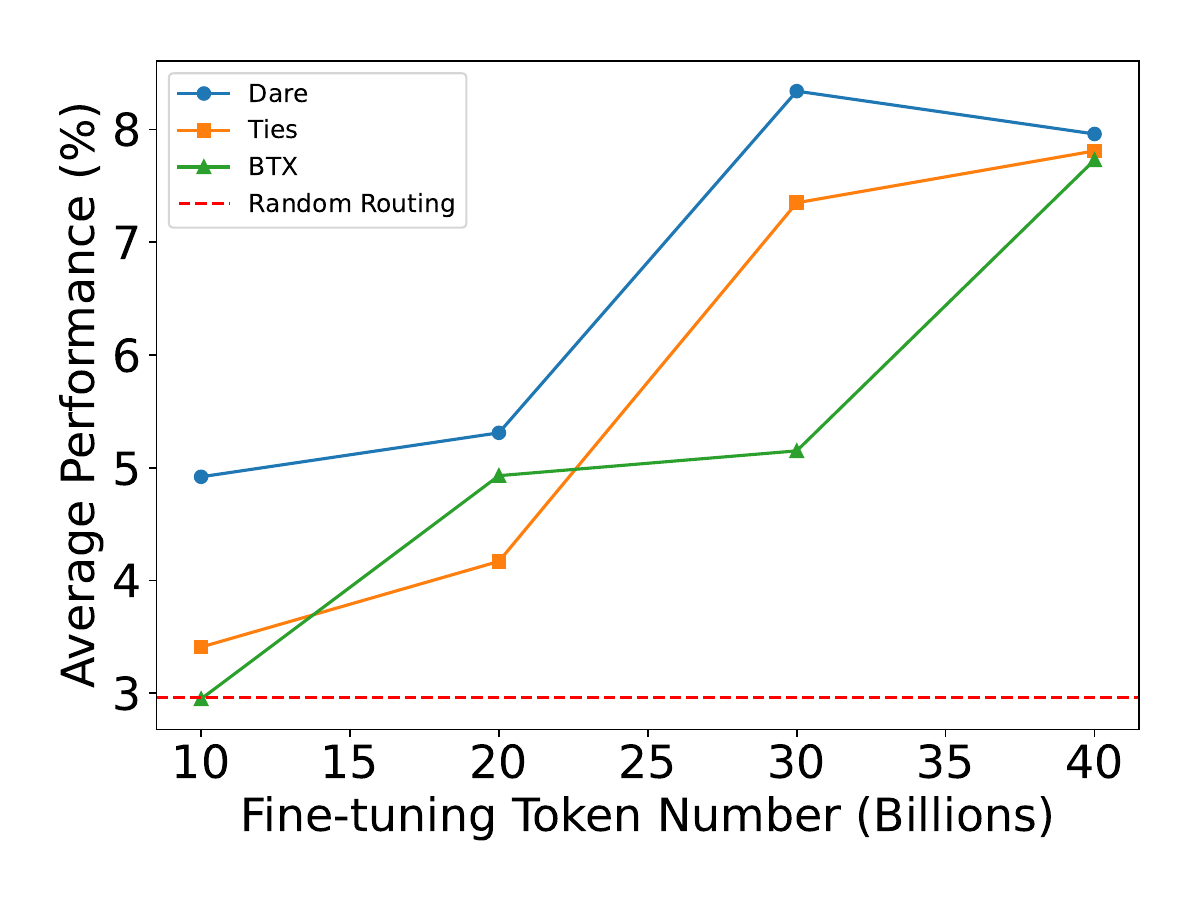}
        \caption{GSM8K}
    \end{subfigure}
    \begin{subfigure}[b]{0.48\columnwidth}
        \centering
        \includegraphics[width=\textwidth]{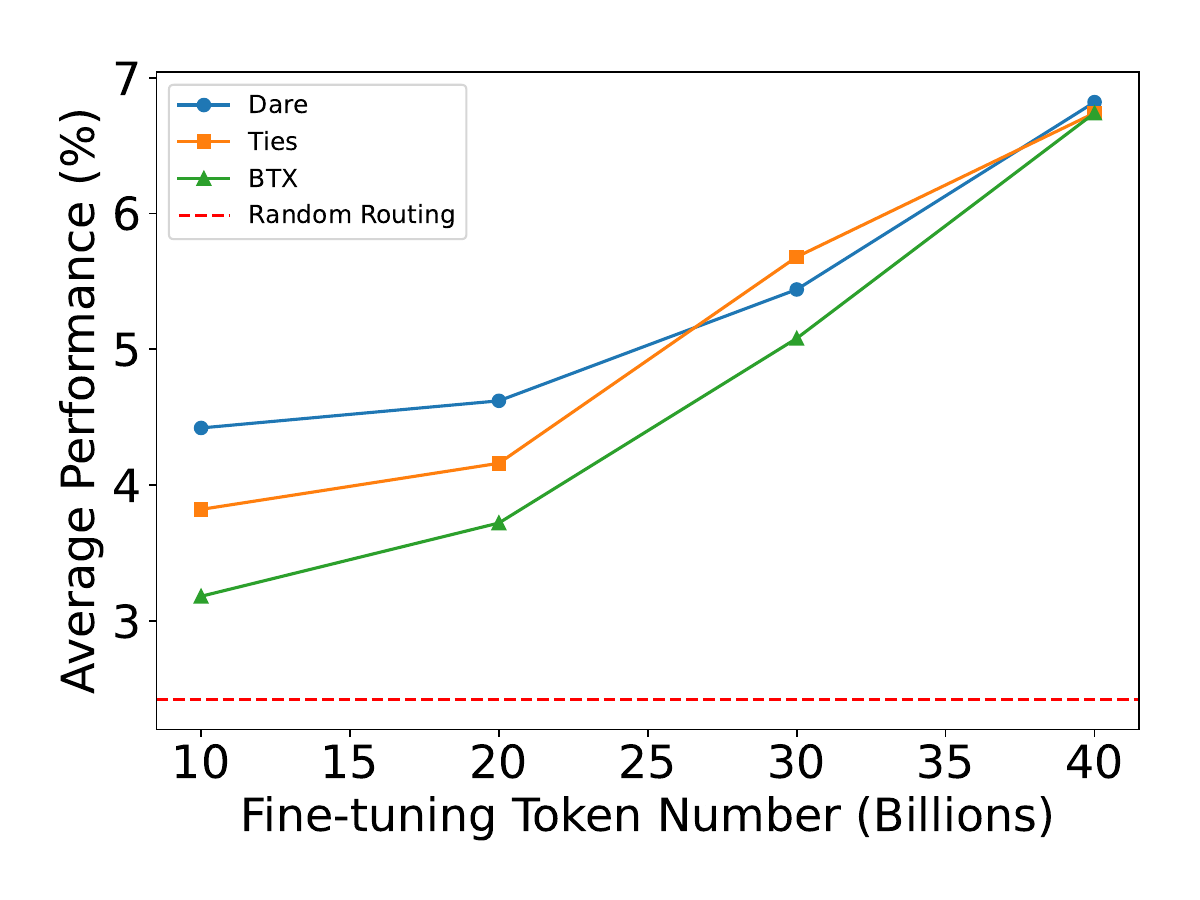}
        \caption{MATH}
    \end{subfigure}
    \begin{subfigure}[b]{0.48\columnwidth}
        \centering
        \includegraphics[width=\textwidth]{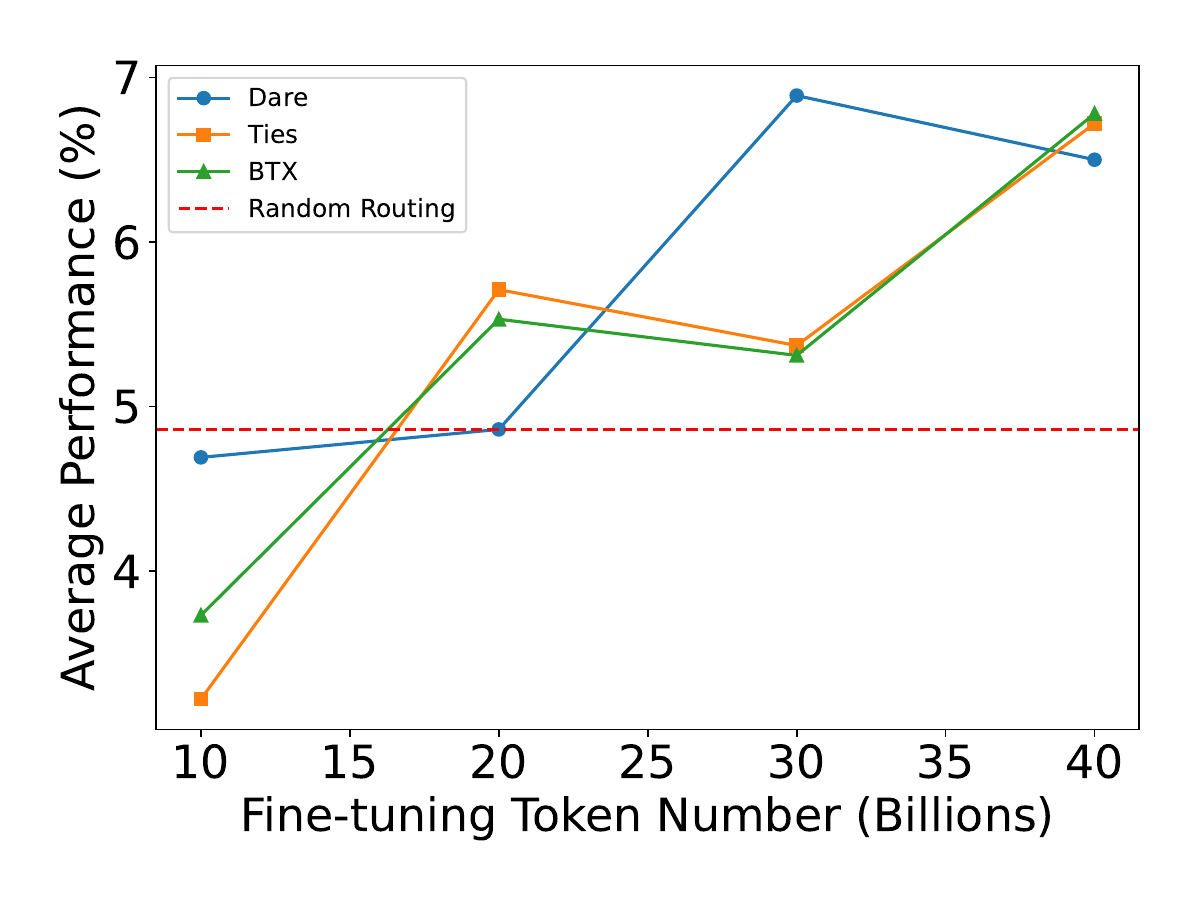}
        \caption{Natural Questions}
    \end{subfigure}
    \begin{subfigure}[b]{0.48\columnwidth}
        \centering
        \includegraphics[width=\textwidth]{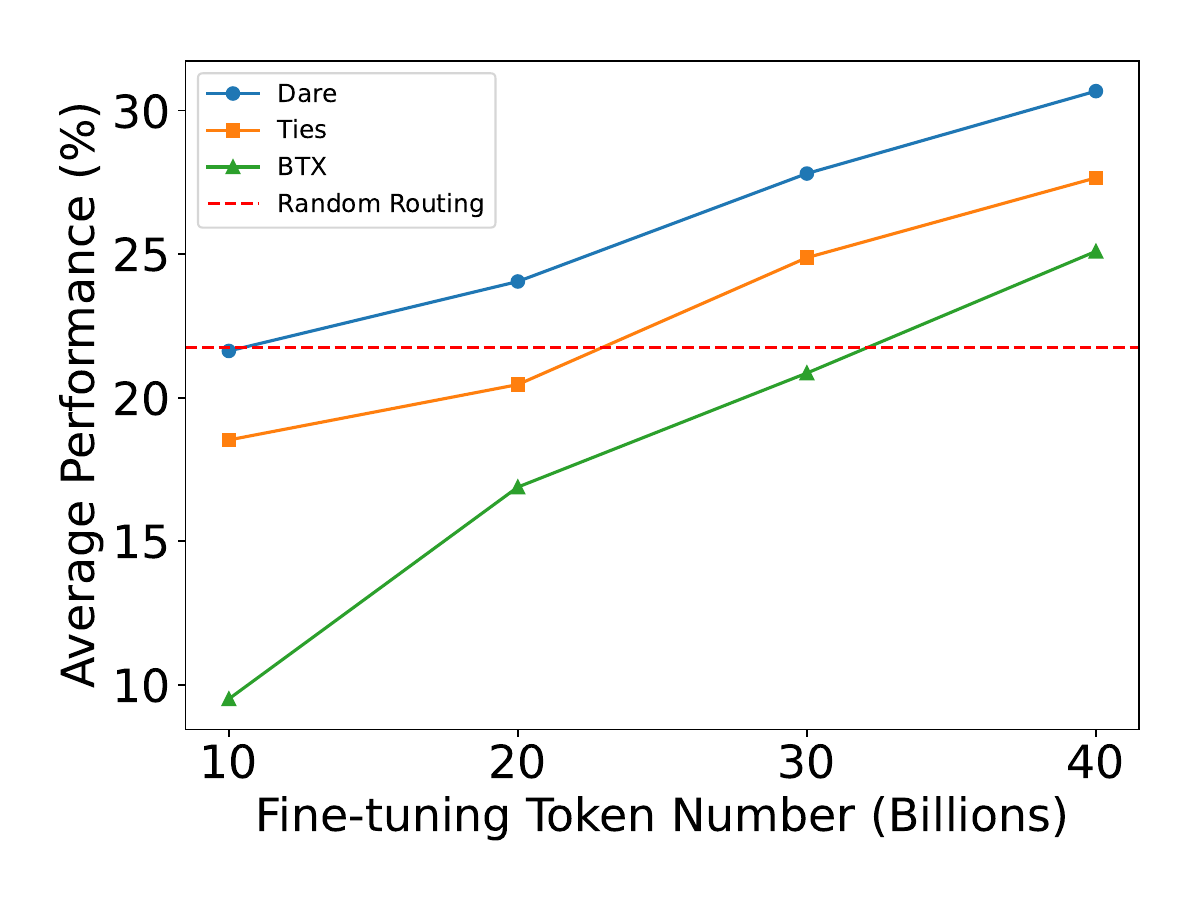}
        \caption{TriviaQA}
    \end{subfigure}
    \caption{Performance with varied fine-tuning token numbers across different datasets.}
    \label{fig:result_token}
\end{figure}


For the calculation of training cost for each method, we will use the product of the number of model parameters and the number of training tokens as a metric for training cost. We present the training costs for each method featured in Tables \ref{table:homo_results}, \ref{table:moe_wo_finetune}, and \ref{table:hetero_results}.

\begin{table}[!htb]
\centering
\resizebox{\columnwidth}{!}{%
\begin{tabular}{lc}
\toprule
\textbf{Method} & \textbf{Training Cost (\# B parameters × \# B tokens)} \\
\midrule
Base-1B & 0 \\
Code Expert & 100 \\
Math Expert & 100 \\
Knowledge Expert & 100 \\
Random Routing & 300 \\
Router Fine-Tuning & 300 \\
BTX Merging & 448 (3 × 100 + 3.7 × 40) \\
Ties Merging & 448 \\
Dare Merging & 448 \\
Model Upcycling & 1258 (3.7 × 340) \\
\bottomrule
\end{tabular}
}
\caption{\label{table:1_training_cost} Training cost of methods in Table \ref{table:homo_results}}
\end{table}

\begin{table}[!htb]
\centering
\resizebox{\columnwidth}{!}{%
\begin{tabular}{lc}
\toprule
\textbf{Method} & \textbf{Training Cost (\# B parameters × \# B tokens)} \\
\midrule
Dare & 100 \\
Ties & 100 \\
Merge Attention & 100 \\
Separate Attention & 100 \\
\bottomrule
\end{tabular}
}
\caption{\label{table:3_training_cost} Training cost of methods in Table \ref{table:moe_wo_finetune}}
\end{table}

\begin{table}[!htb]
\centering
\resizebox{\columnwidth}{!}{%
\begin{tabular}{lc}
\toprule
\textbf{Method} & \textbf{Training Cost (\# B parameters × \# B tokens)} \\
\midrule
Base-1B & 0 \\
Base TinyLlama & 0 \\
Base Olmo & 0 \\
Code Expert & 100 \\
Math TinyLlama & 100 \\
Math Olmo & 100 \\
Knowledge Expert & 100 \\
3-expert MoE & 312 (2 × 100 + 2.8 × 40) \\
(Ours) MoE w/ Math Olmo & 448 \\
(Ours) MoE w/ Math TinyLlama & 448 \\
\bottomrule
\end{tabular}
}
\caption{\label{table:4_training_cost} Training cost of methods in Table \ref{table:hetero_results}}
\end{table}

\end{document}